\documentclass{article}
\usepackage{arxiv}

\usepackage[T1]{fontenc}
\usepackage[latin1]{inputenc}
\usepackage{mathtools,amsthm,amssymb,bbm}
\usepackage[all,warning]{onlyamsmath}
\usepackage{mathtools,bm}
\mathtoolsset{showonlyrefs}
\usepackage{booktabs}
\usepackage{empheq}
\usepackage{fancybox}
\usepackage{minibox}
\usepackage{harmony}
\usepackage{placeins}

\usepackage{xcolor}
\usepackage{graphicx}
\usepackage{subfigure}

\usepackage{natbib}
\usepackage{hyperref}
\usepackage{xcolor}
\hypersetup{
    colorlinks,
    linkcolor={red!50!black},
    citecolor={blue!50!black},
    urlcolor={blue!80!black}
}

\let\cite\undefined

\newcommand{\sref}[2]{\hyperref[#2]{#1~\ref{#2}}}

\input{Autoref.tex}

\newcommand\pythonoutput[1]{\textbf{#1}}
\newcommand\pythonoutputb[1]{\textbf{\textit{#1}}}

\newcommand\thetitle{New Tricks for Estimating Gradients of Expectations}

\newcommand\theabstract{
	We introduce a family of \textit{pairwise stochastic gradient estimators} for gradients of expectations, which are related to the \textit{log-derivative trick}, but involve pairwise interactions between samples. The simplest example of our new estimator, dubbed the \textit{fundamental trick estimator}, is shown to arise from either a) introducing and approximating an integral representation based on the fundamental theorem of calculus, or b) applying the \textit{reparameterisation trick} to an implicit parameterisation under infinitesimal perturbation of the parameters. From the former perspective we generalise to a reproducing kernel Hilbert space representation, giving rise to a locality parameter in the pairwise interactions mentioned above, yielding our \textit{representer trick estimator}. The resulting estimators are unbiased and shown to offer an independent component of useful information in comparison with the log-derivative estimator. We provide a further novel theoretical analysis which further characterises the variance reduction afforded by the new techniques. Promising analytical and numerical examples confirm the theory and intuitions behind the new estimators.
}

\usepackage{pifont}%
\usepackage{amsmath}

\newcommand{\xl}{x_\mathrm{l}}
\newcommand{\xr}{x_\mathrm{r}}

\newcommand{\cmark}{\ding{51}}%
\newcommand{\xmark}{\ding{55}}%
\newcommand{\ww}{w(\mathcal H,f,\bm x,\bm z)}

\DeclareMathOperator*{\cov}{\mathbb Cov}
\DeclareMathOperator*{\corrop}{\mathbb Corr}
\DeclareMathOperator{\varop}{\mathbb V}
\DeclareMathOperator{\expectop}{\mathbb E}

\DeclareMathOperator*{\sign}{sign}
\DeclareMathOperator*{\laplace}{Laplace}

\newcommand{\eat}[1]{}
\newcommand*\intd{\mathop{}\!\mathrm{d}}
\newcommand\expect[2]{\expectop_{#1}\big[#2\big]}
\newcommand\expectlr[2]{\expectop_{#1}\left[#2\right]}
\newcommand\var[1]{\varop\big[#1\big]}
\newcommand\corr[2]{\corrop\big[#1, #2\big]}
\newcommand\varlr[1]{\varop\left[#1\right]}
\newcommand\varwrt[2]{\varop_{#1}\big[#2\big]}
\newcommand\varwrtlr[2]{\varop_{#1}\left[#2\right]}

\newcommand\mo{^{-1}}

\newcommand\at[2]{\left.#1\right|_{#2}}

\newcommand\half{\frac{1}{2}}

\newcommand\ie{\textit{i.e.}}

\newcommand\lcb{\left\{}
\newcommand\rcb{\right\}}
\newcommand\lnb{\left(}
\newcommand\rnb{\right)}
\newcommand\lsb{\left[}
\newcommand\rsb{\right]}
\newcommand\abs[1]{\left|#1\right|}
\newcommand\norm[1]{\left\|#1\right\|}

\newcommand{\realset}{\mathbb{R}}
\newcommand{\ttran}{^\top}

\newcommand\theoremname[1]{(#1)}
\newcommand\prooflabel[1]{(Of #1.)}

\newcommand\rkhs{r.k.h.s.}

\newtheorem{theorem}{Theorem}[section]
\newtheorem{corollary}{Corollary}[theorem]
\newtheorem{lemma}[theorem]{Lemma}

\newtheorem{definition}{Definition}[section]

\begin{document}

\newcommand{\dsosymbol}{{\ensuremath $^\natural$}}
\newcommand{\anusymbol}{{\ensuremath $^\S$}}
\newcommand{\ecolesymbol}{{\ensuremath $^\&$}}
\newcommand{\sydneysymbol}{{\ensuremath $^\flat$}}
\newcommand{\japansymbol}{{\ensuremath $^\sharp$}}
\newcommand{\sssss}{~~}

\title{\thetitle}
\author{%
Christian Walder~\dsosymbol\anusymbol
  \and
Paul Roussel ~\ecolesymbol\dsosymbol
\and
 Cheng Soon Ong~\dsosymbol\anusymbol
\and
Richard Nock~\dsosymbol\anusymbol\sydneysymbol
\hspace{15mm}
Masashi Sugiyama~\japansymbol \thanks{Masashi was supported by JST CREST JPMJCR18A2.}
\\
~ \\
\dsosymbol CSIRO Data61 \sssss \anusymbol Australian National University \sssss \ecolesymbol \'Ecole Polytechnique Paris Saclay \\
\sydneysymbol University of Sydney \sssss \japansymbol RIKEN and University of Tokyo \\ 
~ \\
  \texttt{first.last@\{data61.csiro.au,polytechnique.edu\}, sugi@k.u-tokyo.ac.jp}
}
\date{}

\maketitle

\begin{abstract}
\theabstract
\end{abstract}

\section{Introduction}

Our goal is to estimate the gradient of an expectation of some function $f$ with respect to the parameters of the distribution over which the expectation is taken, that is
\begin{align}
\label{eqn:goal}
\nabla_\theta \expect{\bm x \sim p(\cdot|\theta)}{f(\bm x)}.
\end{align}
This is an important sub-problem of various numerical problems --- in machine learning this includes policy gradient methods of reinforcement learning \citep{reinforce}, training variational auto-encoders \citep{vae}, and variational inference \citep{Jordan:1999:IVM:339248.339252,Blei2016VariationalIA}. 

Sample efficient estimates of \eqref{eqn:goal} are highly desirable as they typically feed into optimisation settings where sampling $\bm x\sim p(\cdot | \theta)$ is expensive \citep{pflugbook,casellabook,pmlr-v32-rezende14,gu2015muprop,SilverHuangEtAl16nature,dice,vadam}. This has inspired a large and growing body of work on variance reduction techniques \citep{NIPS2013_5034,NIPS2017_6961,NIPS2017_7268,tucker2017rebar} ---  %
for an overview see \textit{e.g.} \citet{domkecontrolvariates}, which studies the \textit{control variate} method of reducing variance by linearly combining multiple estimators. 

Despite significant efforts, just two main families of estimators are widely used by the machine learning community \citep{mohamed2019monte}.
The \textbf{reparameterisation trick} \citep{Rubinstein1992,vae,pmlr-v32-rezende14} samples $\bm x\sim p(\cdot|\theta)$ as $\bm \epsilon\sim\mathcal E$ and $\bm x=t_\theta(\bm \epsilon)$, so
\begin{align}
\label{eqn:reparamtrick:sum}
\nabla_\theta \expect{\bm x \sim p(\cdot|\theta)}{f(\bm x)}
 = 
\expect{\bm \epsilon \sim \mathcal E}{\nabla_\theta f(t_\theta(\bm \epsilon))} 
 \approx 
 \frac{1}{n}
 \sum_{i=1}^n
 \nabla f (t_\theta(\bm \epsilon_i))^\top \nabla_\theta t_\theta(\bm \epsilon_i),
\end{align}
where $\bm \epsilon_i\sim \mathcal E$. This simplifies \eqref{eqn:goal} by moving $\theta$ from the distribution to a deterministic function. 
The \textbf{log-derivative trick} %
\citep{miller1967,reinforce} uses $\nabla p  = p \times \nabla \log p $, to obtain
\begin{align}
\nabla_\theta \expect{\bm x \sim p(\cdot|\theta)}{f(\bm x)}
 = 
\expect{
\bm x \sim p(\cdot|\theta)
}{
f(\bm x) \nabla_\theta \log p(\bm x|\theta)
}
 \approx 
 L_n \equiv \frac{1}{n}
 \sum_{i=1}^n f(\bm x_i) \nabla_\theta \log p(\bm x_i|\theta),
\label{eqn:logtrick:sum}
\end{align}
where $\bm x_i\sim p(\cdot|\theta)$. However $L_n$ typically suffers higher variance than the reparameterisation trick \eqref{eqn:reparamtrick:sum}; an observation which is only recently being theoretically explained \citep{reparamtheory}.
\begin{figure}[t]%
\hfill
\begin{minipage}{0.45\textwidth}
  \includegraphics[page=1,width=\textwidth]{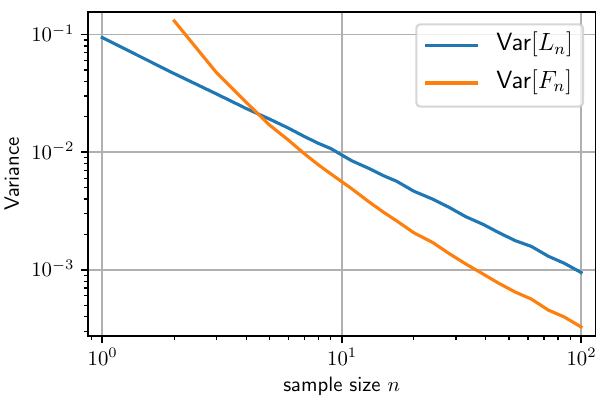}
\end{minipage}
\hfill \hfill \hfill
\begin{minipage}{0.51\textwidth}
  \caption{
  \label{fig:symdemo}
  Variance \textit{vs.} sample size $n$ on the toy problem of \sref{Section}{sec:experiments:analytical}, for the standard log-derivative estimator $L_n$ of \eqref{eqn:logtrick:sum} (blue), and the fundamental trick estimator $F_n$ of \eqref{eqn:logreparameterisationtrick:scalar:doublesum} (orange). Our $F_n$ has favourable (lower) variance for $n \geq 5$. \newline \newline
  } 
\end{minipage}
\hfill 
\end{figure}
\subparagraph{Paper Organisation} 

We provide a new family of \textit{pairwise stochastic gradient estimators} for \eqref{eqn:goal} (see \Autoref{table:ticks} for a comparison), starting with our \textit{fundamental trick estimator} in \Autoref{sec:ft}. We provide theoretical analysis in \Autoref{sec:theoretical}, where we characterise the variance of our estimators, relate the fundamental trick to both the reparameterisation and log-derivative tricks, and show that our techniques yield a new and useful source of information. A shortcoming of the fundamental trick estimator is that the pairwise interactions are crudely non-local. We address this by introducing the more advanced \textit{representer trick estimator} in \Autoref{sec:representer}, which is based on a reproducing kernel Hilbert space (\rkhs ) perspective. This yields a family of estimators parameterised by the choice of \rkhs, and requires certain quantities which we provide in closed form in \Autoref{sec:sobolev} for two important examples of \rkhs\ (namely a particular Sobolev space on compact and non-compact domains). Taking our novel univariate \rkhs\ based schemes as a starting point, we derive multivariate generalisations of the required closed form expressions in \Autoref{sec:multivariate}. Finally, we provide some intuition using analytical and numerical examples in \Autoref{sec:experiments}, before concluding in \Autoref{sec:conclusion}.

\section{A Special Case: The Fundamental Trick Estimator}
\label{sec:ft}

To set the stage, we begin with the simplest special case of our new family of estimators of \eqref{eqn:goal}, the \textit{fundamental trick estimator} for univariate distributions, featured also in \Autoref{fig:symdemo}, 
\begin{empheq}[box=\ovalbox]{align}
\label{eqn:logreparameterisationtrick:scalar:doublesum}
 F_n \equiv 
\frac{1}{n}
\sum_{i=1}^n
\frac{1}{n-1}
\sum_{\substack{j=1 \\ i\neq j}}^n
\frac{f'(x_j)\frac{1}{2}\sign (x_i-x_j) }{p(x_j|\theta)} 
\nabla_\theta \log p(x_i|\theta),
\end{empheq}
where $x_i\sim p(\cdot|\theta)$, and non-bold indicates univariateness.
We have the simple property,
\begin{lemma}
\label{lemma:unbiasedness}
\theoremname{Unbiasedness of $F_n$.}
Let $f:\realset\rightarrow\realset$ be a continuously differentiable function and let $p$ be a density which is absolutely continuous with respect to itself, that is for which $p(x|\theta+h)\rightarrow 0$ for all $h\rightarrow 0_+$ and all points $x$ for which $p(x|\theta)>0$. Then $F_n$ of \eqref{eqn:logreparameterisationtrick:scalar:doublesum} is an unbiased estimator of $\eqref{eqn:goal}$.
\end{lemma}
\begin{proof}
\prooflabel{\sref{Lemma}{lemma:unbiasedness}}
It is sufficient to consider the minimal $n=2$, whereupon $F_2$ contains two terms. Since for any function $g$ and samples $x$ and $z$ drawn i.i.d. from any distribution $\mathcal D$,
\begin{align}
\expectlr{x,z\sim \mathcal D}{g(x,z)} = \expectlr{x,z\sim \mathcal D}{g(z,x)},
\label{eqn:symmetrisepairs}
\end{align}
we need only consider one of these two terms. Writing $x\equiv x_1$ and $z \equiv x_2$, 
\begin{align}
& \expect{x,z%
}{\frac{f'(z) \frac{1}{2} \sign (x-z)}{p(z|\theta)} \nabla_\theta \log p(x|\theta)}
 = 
\iint
p(x|\theta)
 f'(z) 
 \frac{1}{2} \sign (x-z)
\nabla_\theta \log p(x|\theta)
\intd x \intd z
\\ & = 
\nabla_\theta
\iint
p(x|\theta)
f'(z) 
\frac{1}{2} \sign (x-z)
\intd x \intd z
 = 
\nabla_\theta
\expect{x\sim p(\cdot|\theta)}{
\int
f'(z) 
\frac{1}{2} \sign (x-z)
\intd z
}
\\ & = 
\nabla_\theta
\expect{x\sim p(\cdot|\theta)}{f(x)+\text{const.}}
 = 
\label{eqn:comical}
\nabla_\theta
\expect{x\sim p(\cdot|\theta)}{f(x)}.
\end{align}
The second last line uses the fact that the fundamental theorem of calculus \citep{ft} may be expressed 
\begin{align}
f(x)
=
\int_{-\infty}^x + f'(z) \intd z + C_1
 =
\int_x^{+\infty} -f'(z) \intd z + C_2
 = 
\int_{-\infty}^{+\infty}  f'(z) \frac{1}{2} \sign(x-z) \intd z + C,
\label{eqn:symmetriseft}
\end{align}
where the second equality does not follow from the first but may be proven in a similar fashion to it, and final expression is the average of the preceding two.
Note that the requirement $p(x|\theta>0)$ stated under approprate conditions within \autoref{lemma:unbiasedness} is one of absolute continuity of $p$ with respect to itself in a certain sense --- see \citep[\S ~7.3.2 and Appendix~B.4]{glasserman2004monte}.
\end{proof}

The proof of \autoref{lemma:unbiasedness} suggests an interpretation of $F_n$ as $L_n$ with the (known) function $f$ replaced by a noisy Monte Carlo estimate of $f$. However, $F_n$ is not merely equal to $L_n$ plus independent noise.

\begin{table*}[t]
\caption{ 
A qualitative comparison of the Monte Carlo estimators of \eqref{eqn:goal} considered in this paper.
\label{table:ticks}
}
\vspace{3mm}
\centering
{
\footnotesize
\begin{tabular}{c|c|c|c|c}
\toprule
trick: 
 & reparameterisation
 & log-derivative
 & fundamental
 & representer
 \\
 \midrule
reference:
 & \citet{Rubinstein1992} 
 & \citet{miller1967} 
 & (this paper)
 & (this paper)
 \\
equation: 
 & \eqref{eqn:reparamtrick:sum}
 & \eqref{eqn:logtrick:sum}
 & \eqref{eqn:logreparameterisationtrick:scalar:doublesum}
 & \eqref{eqn:pairwisevariance:mn}
 \\
 \midrule
uses $f$ &  \xmark & \cmark & \xmark & \cmark \\
uses $\nabla f$ & \cmark & \xmark & \cmark & \cmark \\
uses $\nabla \log p$ & - & \cmark & \cmark & \cmark \\
uses $p$ & - & \xmark & \cmark & \cmark \\
reparam. & \cmark & \xmark & \xmark & \xmark \\
pairwise & \xmark & \xmark & \cmark & \cmark \\
 \bottomrule
\end{tabular}
}
\end{table*}

\subparagraph{Local Transformation Derivation}
\label{sec:derivation:intuitive}

We provide intuition for \eqref{eqn:logreparameterisationtrick:scalar:doublesum} by deriving it constructively, by applying both the log-derivative \textit{and} reparameterisation tricks --- see \Autoref{fig:cartoon} on page \pageref{fig:cartoon} of the appendix.
 These estimators have been combined in a different and powerful manner by \citet{NIPS2016_6328}, and more recently in a variety of other ways, all rather distinct to the present idea \citep{DBLP:journals/corr/MaddisonMT16,tucker2017rebar,JanGuPoo17,grathwohl2017backpropagation}.
Our idea is to express infinitesimal changes in $\theta$ by a reparameterisation, and then to apply the chain rule (as done in the reparameterisation trick) to that reparameterisation --- for the full derivation see \sref{Appendix}{sec:gore}.

\newcommand\xoneton{x_1, \dots, x_n}
\newcommand\xtwoton{x_2, \dots, x_n}
\newcommand\xonetonmo{x_1, \dots, x_{n-1}}
\newcommand\xonetoi{x_1, \dots, x_i}
\newcommand\xonetoimo{x_1, \dots, x_{i-1}}
\newcommand\xmi{x_1, \dots, x_{i-1}, x_{i+1}, \dots, x_n}
\newcommand\xonetoimtwo{x_1, \dots, x_{i-2}}
\newcommand\xiton{x_{i}, \dots, x_n}
\newcommand\xipoton{x_{i+1}, \dots, x_n}

\section{The General Case: Pairwise Stochastic Gradient Estimators}
\label{sec:theoretical}
We introduce the family of estimators of which $F_n$ of \eqref{eqn:logreparameterisationtrick:scalar:doublesum} is a special case, and establish some basic properties. Additional proofs are left to \sref{Appendix}{sec:varproofs}. 
The estimators are given by two definitions:
\begin{definition}
	\label{def:mn:f}
	\theoremname{Pairwise representation function.} 
	Let $p(\cdot|\theta)$ be a probability distribution with domain $\mathcal X$ and let $f:\mathcal X\rightarrow \mathbb R$. 
	$G:\mathcal X \times \mathcal X \rightarrow \realset$ 
	is a \textit{representation function for $f$ with respect to $p(\cdot|\theta)$} if 
	\begin{align}
		\expect{z\sim p(\cdot|\theta)}{G(x,z)} = f(x), ~~ \forall x\in \mathcal X.
	\end{align}
\end{definition}
\begin{definition}
	\label{def:mn}
	\theoremname{Pairwise stochastic gradient estimator.}
	Let $x_1, x_2, \dots, x_n$ be i.i.d. random variables with domain $\mathcal X$ and density function $p(\cdot|\theta)$ that satisfies the same absolute continuity propery as \autoref{lemma:unbiasedness}, and let $G$ be a pairwise representation function for some $f$ with respect to $p(\cdot|\theta)$. 
	Then we define 
	\begin{empheq}[box=\ovalbox]{align}
		M_n(G)
		\equiv 
		\frac{1}{n} \sum_{i=1}^n 
		\frac{1}{n-1} \sum_{\substack{j=1 \\ i\neq j}}^n
		G(x_i,x_j)\nabla_\theta \log p(x_i|\theta),
		\label{eqn:pairwisevariance:mn}
	\end{empheq}
	to be the pairwise stochastic gradient estimator for \eqref{eqn:goal}.
\end{definition}
The remainder of this section analyses $M_n(G)$, starting with the first and second moments:
\begin{theorem}
	\label{thm:pairwiseunbiasedness}
	\theoremname{Unbiasedness of \eqref{eqn:pairwisevariance:mn}.}
	$\expect{\xoneton\sim p(\cdot|\theta)}{M_n(G)}=\nabla_\theta \expect{x \sim p(\cdot|\theta)}{f(x)}$.
\end{theorem}
\begin{theorem}
	\label{thm:pairwisevariance}
	\theoremname{Variance of \eqref{eqn:pairwisevariance:mn}.}
	$M_n(G)$ has variance
	\begin{align}
		\var{M_n(G)}=\frac{
		\varwrt{x\sim p(\cdot|\theta)}{f(x)\nabla_\theta \log p(x|\theta)+\expectlr{z\sim p(\cdot|\theta)}{G(z,x)\nabla_\theta \log p(z|\theta)}}
		}{n}+\mathcal O(\frac{1}{n^2}). ~~~~~~~
		\label{eqn:pairwisevariance}
	\end{align}
\end{theorem}
It is instructive to apply \sref{Theorem}{thm:pairwisevariance} (proven at the end of this section) to the fundamental trick.
\begin{definition}
	\theoremname{Fundamental trick pairwise representation function.}
	We define $G_\mathrm{ft}: \realset\times\realset\rightarrow\realset$,
	\begin{align}
		G_\mathrm{ft}(x,y) = \frac{\frac{1}{2} f'(x)\sign(z-x)}{p(x|\theta)}.
		\label{eqn:mn:f:ft}
	\end{align}
\end{definition}
If is easy to check that $G_\mathrm{ft}$ is a pairwise representation function, and that it yields the fundamental trick $F_n$ of \eqref{eqn:logreparameterisationtrick:scalar:doublesum}, \textit{i.e.} $M_n(G_\mathrm{ft})=F_n$. This allows a useful corollary of \sref{Theorem}{thm:pairwisevariance}, namely
\begin{corollary}
	\label{cor:ftvariance}
	\theoremname{Fundamental trick variance.}
    Assume that $p(\cdot|\theta)$  admits a reparameterisation of the standard type (\textit{i.e.} that employed in \eqref{eqn:reparamtrick:sum}). Then with the assumptions of \sref{Theorem}{thm:pairwisevariance}, 
	\begin{align}
		\var{F_n} = \frac{
		\varwrt{x\sim p(\cdot|\theta)}{f'(x)\nabla_\theta \log p(x|\theta)+f'(x)\frac{\intd x}{\intd \theta}}
		}{n}+\mathcal O(\frac{1}{n^2}),
		\label{eqn:ftvariance}
	\end{align}
	where $f'(x)\frac{\intd x}{\intd \theta}$ is the standard reparameterisation gradient, in the notation of \citet{irpgradients}.
\end{corollary}
\begin{proof}
	\prooflabel{\sref{Corollary}{cor:ftvariance}}
	The result follows from \sref{Theorem}{thm:pairwisevariance} by noting that 
	\begin{align}
		& \expectlr{z\sim p(\cdot|\theta)}{G(z,x)\nabla_\theta \log p(z|\theta)}
		 =
		\int G(x,z) \nabla_\theta \log p(z|\theta) p(z|\theta) \intd z
		\\ 
		& =
		\int \frac{\frac{1}{2} f'(x)\sign(z-x)}{p(x|\theta)} \nabla_\theta \log p(z|\theta) p(z|\theta) \intd z
		 =
		\frac{f'(x)}{p(x|\theta)} \nabla_\theta \int \frac{1}{2}\sign(z-x) p(z|\theta) \intd z
		\\ & 
		 =
		\frac{f'(x)}{p(x|\theta)} \Big( - \nabla_\theta F(x|\theta)\Big)
		 =
		f'(x) \frac{\intd x}{\intd \theta},
	\end{align}
where $F(x|\theta)$ is the cumulative distribution function of $x|\theta$, and 
we used $\frac{\intd x}{\intd \theta} = - \frac{\nabla_\theta F(z|\theta)}{p(x|\theta)}$, a key relation featured by \citet{irpgradients} as their Equation 8.
\end{proof}
	This shows that for $n \to \infty$, the fundamental trick has four times the variance of the average of the log derivative trick and the reparameterisation trick estimators, should a reparameterisation be available. This is because the argument of the variance in \eqref{eqn:ftvariance} is equal to twice that average, for a single sample $x$. $F_n$ and more generally $M_n(G)$ are nonetheless relevant however, because they
\begin{enumerate}	
	\item apply in different settings (especially when a reparameterisation is unavailable --- see \Autoref{table:ticks}), 
	\item may be superior to either the log-derivative or the reparameterisation trick estimators in isolation (see \Autoref{fig:symdemo} for an example), and
	\item may reduce the variance of the log-derivative estimator in convex combination with it.
\end{enumerate}	
The second point is obvious from the form of \eqref{eqn:ftvariance}. The third point is more subtle, however, and is the subject of the remainder of this section.
We begin by introducing some well known concepts from statistics (see \textit{e.g.} \citet{domkecontrolvariates} for a review tailored to the present context) and financial portfolio theory (see \textit{e.g.} \citet{markowitz}).
\begin{definition}
\theoremname{Minimum variance convex combination.} 
The min-var weights of $m$ random variables $x_1, x_2, \dots, x_m$ is the set of vectors of coefficients $\bm c \in \mathbb R^m$ satisfying $\bm c \ttran \bm 1 = 1\,\,$ for which $\var{\sum_{i=1}^m c_i x_i}$ are minimised.
\end{definition}
The following result is well known, and easily shown with \textit{e.g.} the use of Lagrange multipliers.
\begin{lemma}
\theoremname{Expression for the min-var weights.}
If the min-var weights of the previous definition exist are unique, then they are given by $\bm c^\star \equiv \Sigma\mo \bm 1 / (\bm 1 \ttran \Sigma\mo \bm 1)$, where $\Sigma$ is the covariance matrix given by $\Sigma_{i,j} \equiv \cov(x_i,x_j)$.
\end{lemma}
The crucial property we will verify is that the fundamental trick estimator adds useful information in convex combination with the log-derivative estimator. The notion of adding useful information is formalised by our definition of an important concept in financial portfolio theory, namely
\begin{definition}
\theoremname{Diversification.}
$x_2$ diversifies $x_1$ if $(1, 0)^\top$ is not an associated min-var weight.
\end{definition}
A new estimator is useful in the sense of variance reduction, if it diversifies an existing estimator. Checking whether one estimator diversifies another turns out to be particularly simple if they are unbiased, with the condition involving only the uncentered second moments. In particular we have
\begin{lemma}
\label{lemma:momentcondition}
\theoremname{Uncentered second moment condition for diversification.}
Let $x_1$ and $x_2$ have the same mean and finite second moments. If $\mathbb E\lsb x_1 x_2\rsb \neq \mathbb E\lsb x_1 x_1\rsb$, then $x_2$ diversifies $x_1$.
\end{lemma}
\begin{proof}
\prooflabel{\sref{Lemma}{lemma:momentcondition}}
The min-var weights are given by $\bm c^\star \propto ( \Sigma_{22}- \Sigma_{12},  \Sigma_{11}- \Sigma_{12})\ttran$. Since $\expect{}{x_1}=\expect{}{x_2}\equiv \mu$, we have $\Sigma_{i,j} = \expect{}{x_i x_j}-\mu^2 \equiv \hat \Sigma_{i,j}-\mu^2$. Hence
\begin{align}
\bm c^\star \propto (\hat \Sigma_{22}-\hat \Sigma_{12}, \hat \Sigma_{11}-\hat \Sigma_{12})\ttran,
\label{eqn:cstar}
\end{align}
and so 
if $\mathbb E\lsb x_1 x_2\rsb \neq \mathbb E\lsb x_1 x_1\rsb$ then the second element of $\bm c^\star$ is non-zero.
\end{proof}
We may now state our next important result, which is non-trivial in the sense that there are closely related estimators which fail to diversify in this way --- see \sref{Appendix}{sec:hsl} for an example.
\begin{theorem}
	\label{thm:diversification:mnln}
	\theoremname{Usefulness of $M_n(G)$ in combination with $L_n$.}
	If both  $L_n$ of \eqref{eqn:logtrick:sum} and $M_n(G)$ of \eqref{eqn:pairwisevariance:mn} have finite variance then in general $M_n(G)$ diversifies $L_n$.
\end{theorem}

\begin{proof}
\prooflabel{\sref{Theorem}{thm:diversification:mnln}}
It is sufficient to consider $L_n$ and $M_n(G)$ with $n=2$, and a single term in the summation for $M_2(G)$. According to \sref{Lemma}{lemma:momentcondition}, we require that $\Sigma_\mathrm{ll}\neq\Sigma_\mathrm{lm}$, where
\begin{align}
    \Sigma_\mathrm{ll}
    & = 
    \mathbb E_{\bm x, \bm z} \left[\left(\half\big(
    f(\bm x)\nabla_\theta\log p(\bm x|\theta)
    +
    f(\bm z)\nabla_\theta\log p(\bm z|\theta)
    \big)\right)^2\right]
    \\
    & = 
    \half \, 
    \mathbb E_{\bm x} \left[
    \big(f(\bm x)\nabla_\theta\log p(\bm x|\theta)\big)^2
    \right]
    +
    \half \,
    \underbrace{
    \mathbb E_{\bm x} \left[
    f(\bm x)\nabla_\theta\log p(\bm x|\theta)
    \right]^2
    }_{\equiv \nu_\mathrm{ll}},
\shortintertext{and}
    \Sigma_\mathrm{lm}
    & = 
    \mathbb E_{\bm x, \bm z} \left[\left(\half\big(
    f(\bm x)\nabla_\theta\log p(\bm x|\theta)
    +
    f(\bm z)\nabla_\theta\log p(\bm z|\theta)
    \big)\right)
    G(\bm x, \bm z)
\nabla_\theta \log p(\bm x|\theta)
\right]
    \\
    & = 
    \half \, 
    \mathbb E_{\bm x} \Big[
    f(\bm x)\nabla_\theta\log p(\bm x|\theta)E_{\bm z} \big[ G(x,z) \big]\nabla_\theta\log p(\bm x|\theta)
    \Big]
    +
    \half \,
    \underbrace{
    \mathbb E_{\bm x, \bm z} \left[
    f(\bm z)\nabla_\theta\log p(\bm z|\theta)
	G(\bm x, \bm z)
	\nabla_\theta \log p(\bm x|\theta)
    \right]
    }_{\equiv \nu_\mathrm{lm}}
    \\
    & = 
    \half \, 
    \mathbb E_{\bm x} \left[
    \big(f(\bm x)\nabla_\theta\log p(\bm x|\theta)\big)^2
    \right]
    +
    \half \, \nu_\mathrm{lm}.
\end{align}
But the two quantities which differ in these expressions are given by
\begin{align}
    \nu_\mathrm{ll}
     & =
    \int f(\bm x) \nabla_\theta\log p(\bm x|\theta) p(\bm x|\theta) \intd \bm x \int f(\bm z)  \nabla_\theta\log p(\bm z|\theta)  p(\bm z|\theta) \intd \bm z,
    \shortintertext{and}
    \nu_\mathrm{lm}
     & =
    \iint \Big( f(\bm z)\nabla_\theta\log p(\bm z|\theta)
	G(\bm x, \bm z)
	\nabla_\theta \log p(\bm x|\theta)\Big) p(\bm x|\theta) p(\bm z|\theta) \intd \bm x \intd \bm z,
\end{align}
the latter of which is not a separable double integral, hence in general $\nu_\mathrm{ll}\neq\nu_\mathrm{lm}$.
\end{proof}
Our next two results characterise the degree of variance reduction afforded by this diversification.
\begin{theorem}
	\label{thm:combination:mnln}
	\theoremname{Convex combination of log-derivative and pairwise estimators.}	
	For the convex combination of $L_n$ of \eqref{eqn:pairwisevariance:mn} and $M_n(G)$ of \eqref{eqn:logtrick:sum}, we have
	\begin{align}
		& \varlr{\alpha M_n(G) + (1-\alpha)L_n}
		\\ & = 
		\frac{
		\varwrt{x\sim p(\cdot|\theta)}{f(x)\nabla_\theta \log p(x|\theta)+\alpha \expectlr{z\sim p(\cdot|\theta)}{G(z,x)\nabla_\theta \log p(z|\theta)}}
		}{n}+\mathcal O(\frac{\alpha^2}{n^2}).
		\label{eqn:combination:mnln:variance}
	\end{align}
\end{theorem}
\begin{theorem}
	\label{thm:varreduction}
	\theoremname{Optimal variance reduction of $M_n(G)$ in convex combination with $L_n$.}
	Let 
	\begin{align}
		\rho 
		& =
		\corr{f(x)\nabla_x \log p(x|\theta)}{\expect{z\sim p(\cdot|\theta)}{G(x,z)\nabla_\theta \log p(z_\theta)}},
		\shortintertext{and}
		\sigma_L^2 & =
		\varwrt{x\sim p(\cdot|\theta)}{f(x)\nabla_x \log p(x|\theta)}.
	\end{align}
	Then the optimal reduction, relative to the variance of $L_n$, in the $\mathcal O(\frac{1}{n})$ component of the variance afforded by taking a convex combination of $M_n(G)$ with $L_n$, is
	\begin{align}
		\min_{\alpha \in \realset} 
		\frac{ 
		\varwrt{x\sim p(\cdot|\theta)}{f(x)\nabla_\theta \log p(x|\theta)+\alpha \expectlr{z\sim p(\cdot|\theta)}{G(z,x)\nabla_\theta \log p(z|\theta)}}
		}{\sigma_L} = 1-\rho^2.
	\end{align}
\end{theorem}
We conclude by proving \sref{Theorem}{thm:pairwisevariance} using the following result, itself proven in \sref{Appendix}{sec:varproofs}.
\begin{lemma}
	\label{lemma:eve:functional:inverted}
	\theoremname{Inverted functional Eve's law.}
	Let $\xoneton$ be i.i.d. random variables with domain $\mathcal X$, and let $f:\mathcal X^n\rightarrow \realset$ be a deterministic function with $\varwrt{\xoneton}{f(\xoneton)} < \infty$. Then
	\begin{align}
		& \varwrtlr{\xoneton}{f(\xoneton)}
		\\ & = 
		\sum_{i=2}^n \expectlr{x_i}{\varwrtlr{\xonetoimo}{\expectlr{\xipoton}{f(\xoneton)}-\expectlr{\xiton}{f(\xoneton)}}}
		\label{eqn:eve:functional:inverted:a}
		\\ & ~~~~ +
		\sum_{i=1}^n \varwrtlr{x_i}{\expectlr{\xmi}{f(\xoneton)}}.
		\label{eqn:eve:functional:inverted:b}
	\end{align}
\end{lemma}
\begin{proof}
	\prooflabel{\sref{Theorem}{thm:pairwisevariance}}
	The idea is to equate $f(\xoneton)$ of \sref{Lemma}{lemma:eve:functional:inverted} to $M_n(G)$ of \sref{Theorem}{thm:pairwisevariance}, and show that \eqref{eqn:eve:functional:inverted:a} and \eqref{eqn:eve:functional:inverted:b} lead to the $\mathcal O(\frac{1}{n^2})$ and $\mathcal O(\frac{1}{n})$ terms in \eqref{eqn:pairwisevariance}, respectively.

	We first treat the summation on line \eqref{eqn:eve:functional:inverted:b}. Substituting $M_n(G)$ of \eqref{eqn:pairwisevariance:mn} for $f(\xoneton)$, and denoting $\expect{-i}{\cdot} =\expect{\xmi}{\cdot}$, the summand becomes
	\begin{align}
		 & \varwrtlr{x_i}{\expectlr{-i}{M_n}}
		 \\ & \stackrel{(a)}{=}
		 \varwrt{x_i}{\expect{-i}{
		 \frac{1}{n} \sum_{j=1}^n 
		\frac{1}{n-1} \sum_{\substack{k=1 \\ j\neq k}}^n
		G(x_j,x_k)\nabla_\theta \log p(x_j|\theta)
		}}
		 \\ & \stackrel{(b)}{=}
		 \varwrt{x_i}{\expect{-i}{
		 \frac{1}{n(n-1)} 
		 \sum_{\substack{j=1 \\ i\neq j}}^n
		\lnb 
		G(x_j,x_i)\nabla_\theta \log p(x_i|\theta)
		+
		G(x_i,x_j)\nabla_\theta \log p(x_j|\theta)
		\rnb
		}}
		 \\ & \stackrel{(c)}{=}
		 \varwrt{x}{\expect{z}{
		 \frac{1}{n}
		\lnb 
		G(x,z)\nabla_\theta \log p(x|\theta)
		+
		G(z,x)\nabla_\theta \log p(z|\theta)
		\rnb
		}}
		 \\ & \stackrel{(d)}{=}
		\varwrt{x}{
		\frac{1}{n}
		\lnb 
		f(x) \nabla_\theta \log p(x|\theta)
		+
		\expect{z}{G(z,x)\nabla_\theta \log p(z|\theta)
		\rnb
		}}
		 \\ & \stackrel{(e)}{=}
		\frac{1}{n^2}
		\varwrt{x}{
		f(x) \nabla_\theta \log p(x|\theta)
		+
		\expect{z}{G(z,x)\nabla_\theta \log p(z|\theta)
		}},	
		\label{eqn:varsummand:diag}
		\end{align}
where $(a)$ is by definition, $(b)$ uses the fact that terms with $i \neq j$ and $i\neq k$ may be dropped as they are constant inside the variance w.r.t. $x_i$, $(c)$ introduces $x,z\sim p(\cdot|\theta)$ and exploits the i.i.d. assumption on $\xoneton$, $(d)$ is by the definition of $G(\cdot, \cdot)$, and $(e)$ moves a constant outside the variance. 
Hence the summation on line \eqref{eqn:eve:functional:inverted:b} reduces to
\begin{align}
	& \sum_{i=1}^n \varwrtlr{x_i}{\expectlr{\xmi}{f(\xoneton)}}
	\\ & =
	\frac{
		\varwrt{x}{
		f(x) \nabla_\theta \log p(x|\theta)
		+
		\expect{z}{G(z,x)\nabla_\theta \log p(z|\theta)
		}}}{n},
\end{align}
which matches the first term on the r.h.s. of \eqref{eqn:pairwisevariance}.
Finally, we treat the summation on line \eqref{eqn:eve:functional:inverted:a}. Substituting $M_n(G)$ of \eqref{eqn:pairwisevariance:mn} for $f(\xoneton)$, the inner expectation in the summand becomes
\begin{align}
	& \expectlr{x_i}{\varwrtlr{\xonetoimo}{\expectlr{\xipoton}{f(\xoneton)}-\expectlr{\xiton}{f(\xoneton)}}}
	\\ & \stackrel{(a)}{=}
	\expectlr{x_i}{\varwrtlr{\xonetoimo}{\expectlr{\xipoton}{M_n(G)}-\expectlr{\xiton}{M_n(G)}}}
	\\ & \stackrel{(b)}{=}
	\expectlr{x_i}{\varwrtlr{\xonetoimo}{\expectlr{\xipoton}{M_n(G)-\expectlr{x_i}{M_n(G)}}}}
	\\ & \stackrel{(c)}{\equiv}
    \expect{x_i}{\varwrt{\xonetoimo}{\frac{1}{n(n-1)}\expect{\xipoton}{\sum_{\substack{j=1\\i\neq j}}^n \lnb T_{i,j}-\expect{x_i}{T_{i,j}}\rnb}}}
	\\ & \stackrel{(d)}{=}
    \expect{x_i}{\varwrt{\xonetoimo}{\frac{1}{n(n-1)}\sum_{\substack{j=1\\i\neq j}}^n\expect{\xipoton}{ T_{i,j}-\expect{x_i}{T_{i,j}}}}}
	\\ & \stackrel{(e)}{=}
    \expect{x_i}{\varwrt{\xonetoimo}{\frac{1}{n(n-1)}\sum_{j=1}^{i-1}\expect{\xipoton}{ T_{i,j}-\expect{x_i}{T_{i,j}}}}}
	\\ & \stackrel{(f)}{=}
    \expect{x_i}{\varwrt{\xonetoimo}{\frac{1}{n(n-1)}\sum_{j=1}^{i-1}\lnb T_{i,j}-\expect{x_i}{T_{i,j}}\rnb}}
	\\ & \stackrel{(g)}{=}
    \frac{1}{n^2(n-1)^2} \expect{x_i}{\varwrt{\xonetoimo}{\sum_{j=1}^{i-1}\lnb T_{i,j}-\expect{x_i}{T_{i,j}}\rnb}}
	\\ & \stackrel{(h)}{=}
    \frac{(i-1)}{n^2(n-1)^2} \expect{x_i}{\varwrt{x_i,x_j}{\lnb T_{i,j}-\expect{x_i}{T_{i,j}}\rnb}}
	\\ & \stackrel{(i)}{=}
    \frac{(i-1)}{n^2(n-1)^2} \cdot C,
    \label{eqn:varsummand}
 \end{align}
 where ($a$) substitutes $M_n(G)$, ($b$) is by linearity of expectation, ($c$) uses the definition of $M_n(G)$ with $T_{i,j}=G(x_i, x_j)\nabla_\theta \log p(x_i|\theta)+G(x_j, x_i)\nabla_\theta \log p(x_j|\theta)$, ($d$) is again by linearity, ($e$) removes terms with $j\geq i$ as they are constant under the variance, ($f$) removes the expectation w.r.t. variables not present in the argument of the expectation, ($g$) moves a constant out of the variance and expectation, ($h$) applies the Bienaym\'{e} formula for the variance of the sum of independent random variables, and ($i$) introduces a constant $C$ which is independent of $n$ and $i$.
 Putting \eqref{eqn:varsummand} into \eqref{eqn:eve:functional:inverted:a} yields 
 \begin{align}
 	& \sum_{i=2}^n \expectlr{x_i}{\varwrtlr{\xonetoimo}{\expectlr{\xipoton}{f(\xoneton)}-\expectlr{\xiton}{f(\xoneton)}}}
 	\\ & = \sum_{i=2}^n \frac{(i-1)}{n^2(n-1)^2} \cdot C
 	= \frac{C}{n^2(n-1)^2} \sum_{i=2}^n (i-1)
 	= \mathcal O(\frac{1}{n^2}),
 \end{align}
 which matches the second term on the r.h.s. of \eqref{eqn:pairwisevariance} and completes the proof.
\end{proof}

\section{Representer Trick Estimator}
\label{sec:representer}
We now introduce a more powerful version of the fundamental trick  introduced by way of introduction in \Autoref{sec:ft} and as part of the more general framework as \sref{Definition}{def:mn}. 
Intuitively, the $\sign(x_i-x_j)$ in the fundamental trick \eqref{eqn:mn:f:ft} reflects the direction from $x_i$ to $x_j$, and serves as a stochastic proxy for the factor $\nabla f (t_\theta(\epsilon_i))$ in the reparameterisation trick \eqref{eqn:reparamtrick:sum} without needing an explicit reparameterisation (see the supplementary \Autoref{fig:cartoon} for an illustration). 
The non-locality of this $\sign$ term (which ignores the magnitude of $(x_i-x_j)$) is undesirable however, both intuitively and as empirically demonstrated in the supplementary \Autoref{fig:cov_demo_extra}. We  address this by introducing a length scale parameter by way of a particular \rkhs . This effectively weights the contribution of pairs of samples according to the distance between them, in a precise way which retains the properties of \Autoref{sec:theoretical}:

\begin{definition}
	\label{def:mn:f:rkhs}
	\theoremname{The r.k.h.s. pairwise representation function.}
	Let $\mathcal H$ be an r.k.h.s. with domain $\mathcal X$, reproducing kernel $k$ and regularisation operator $P$, so that for $f \in \mathcal H$ and $\forall \bm x \in \mathcal X$, we have \citep{wahbabook,learningwithkernels,scattered,agnanbook}
	\begin{align}
\label{eqn:reproducing}
\smash[t]{
f(\bm x) 
 = 
\langle f, k(\cdot, \bm x) \rangle_\mathcal{H}
 \equiv
\int_{\mathcal X} Pf(\bm z) Pk(\bm z,\bm x) \intd \bm z
 \equiv 
\int_{\mathcal X} \ww \intd \bm z.
}
\end{align}
Further let $p(\cdot|\theta)$ be a probability distribution on $\mathcal X$. Then we define
\begin{empheq}[box=\ovalbox]{align}
	F_{\mathcal H}(x,z) \equiv \frac{\ww}{p(z|\theta)},
	\label{eqn:mn:f:fh}
\end{empheq}
where for clarity we neglect to notate dependence on $f$ and $p(\cdot | \theta)$.
\end{definition}
The point is that it is trivial to show that 
\begin{lemma}
	\label{lemma:mn:h:fh:representation}
	$F_{\mathcal H}(\bm x,\bm  z)$ is a pairwise representation function for $p(\cdot|\theta)$ and $f$.
\end{lemma}

The condition in \sref{Definition}{def:mn:f:rkhs} that the support of $p(\cdot | \theta)$ match the domain of $\mathcal H$  motivates the explicit treatment of r.k.h.s.'s on compact domains in the following \Autoref{sec:sobolev}. %

\subsection{First-Order Sobolev Norm \rkhs}
\label{sec:sobolev}

We now derive closed form examples of the key quantity $w$ of \eqref{eqn:reproducing}. Let $\mathcal H_a(\mathcal X)$ be the \rkhs\ with regularisation operator $P = \mathbbm 1 + a \partial$, $a>0$ on domain $\mathcal X$, where $\partial$ is defined by $\partial f \equiv f'$. $\mathcal H_a(\mathcal X)$ is a Sobolev space with length-scale $a$ \citep{wahbabook}. This choice of $P$ is convenient as it involves only the first derivative, thereby leading to a family of estimators that require only $f$ and $f'$. 

\subsubsection{Case of the whole real line}

On the domain $\mathcal X = \mathbb R$ the operator $\partial$ has adjoint $\partial^*=-\partial$, and so we have
\begin{align}
\langle f , g \rangle_{\mathcal{H}_a(\realset)} 
 =
\langle Pf, Pg \rangle_{L_2(\realset)}
 =
\langle f | (\mathbbm 1 + a \partial)^* (\mathbbm 1 + a \partial) | g \rangle_{L_2(\realset)}
 =
\langle f | \mathbbm 1 - (a \partial)^2 | g \rangle_{L_2(\realset)}.
\end{align}
Denoting 
$
\phi_x(z) \equiv k(z,x)
$, 
the reproducing property \eqref{eqn:reproducing}
implies, $\forall x \in \realset$,
\begin{align}
\lnb\mathbbm 1 - (a \partial)^2\rnb\phi_x(\cdot)
& = 
\delta_x(\cdot),
\label{eqn:sobolevde}
\end{align}
where $\delta_x$ is the Dirac distribution centered at $x$.
Here we impose the boundary conditions
$\phi_x'(-\infty) = \phi_x'(\infty) = 0$
to obtain the well known solution (see e.g. \citet{wahbabook})
$\phi_x(z) \equiv k(z,x) 
 = \exp(-\abs{x-z}/a)/(2a)$.
Letting $\partial$ act on the first argument of $k(\cdot,\cdot)$, we obtain
\begin{align}
P k(z,x)
 & =
k(z,x) + a \partial k(z,x)
 =
k(z,x) + a \sign(x-z)/a \,k(z,x)
\\ & =
(1+\sign(x-z)) \, k(z,x)
 =
2 \mathbb I (z<x) \, k(z,x).
\end{align}
Hence we have the $w$ of \eqref{eqn:reproducing} in closed form,
\begin{align}
w(\mathcal H_a(\realset), f, x, z)
 \equiv 
Pf(z) Pk(z,x)
= 
\lnb f(z) + a f'(z) \rnb 2 \mathbb I (z<x) \, \exp(-\abs{x-z}/a) / (2a), 
\end{align}
however this expression %
may be improved by symmetrisation analagously to \eqref{eqn:symmetriseft}, yielding the following expression for the $w$ of \eqref{eqn:mn:f:fh}, which we verify by first principles in \sref{Appendix}{sec:firstprincipleslaplace},
\begin{empheq}[box=\ovalbox]{align}
\label{eqn:w:sym}
 w_\mathrm{symmetrised}(\mathcal H_a(\realset), f, x, z) \equiv
\lnb f(z) + a \sign (x-z) f'(z)\rnb \exp(-\frac{\abs{x-z}}{a}) / (2a).
\end{empheq}
It is easy to check that putting this into \eqref{eqn:mn:f:fh} and taking $a \to \infty$ yields the fundamental trick \eqref{eqn:mn:f:ft}. Furthermore, in a certain sense the limit $a \to 0$ gives rise to the log-derivative trick estimator.
\subsubsection{Case of a bounded interval}
\label{sec:sobolev:bounded}
For $\mathcal X = [-1, +1] \equiv \mathcal U$ we impose %
$\phi_x'(-1) = \phi_x'(+1) = 0$,
and immediately obtain (using the \textit{Mathematica} software) the result of \citet{agnan1996} and assuming $\xl\leq\xr$ (which is no restriction since reproducing kernels are symmetric functions), we have
\begin{align}
\label{eqn:agnank}
k_a(\xl, \xr) = \lnb \cosh((1+\xl)/a) \cosh((1-\xr)/a)\rnb / \lnb a \sinh(2/a)\rnb.
\end{align}
Letting $\partial$ again act on the first argument of $k(\cdot,\cdot)$, we obtain with some  algebra,
\begin{align}
P k_a(z,x) &
\label{eqn:pkzxcompact}
=\big(1+ \sign (z-x) \tanh\big((1+\sign(z-x) x \big)/a \big)\big) k_a(x,z).
\end{align}

The $w$ of \eqref{eqn:reproducing} for the bounded case is hence
\begin{empheq}[box=\ovalbox]{align}
 w(\mathcal H_a(\mathcal U), f, x, z) \equiv
\label{eqn:w:bounded:closedform}
\big( f(z) + a f'(z) \big) \, Pk_a(z,x), %
\end{empheq}
where $Pk_a(z,x)$ is given in \eqref{eqn:pkzxcompact} and $k_a(z,x)$ in \eqref{eqn:agnank}. As $\phi_x$ is the solution to \eqref{eqn:sobolevde}, we have by definition that $P^* P \phi_x=\delta_x$ on $\mathcal U$. However, the restriction to $\mathcal U$ introduces terms due to the boundary conditions, in particular we have the following representation, which does not appear to have previously been made explicit in the \rkhs\ literature. 

\begin{lemma}
\label{lemma:compactintegralform:main}
Let $f:[-1,+1]\rightarrow\realset$ be continuously differentiable and $a>0$. Then $\forall x \in [-1,+1]$,
\begin{align}
\label{eqn:compactintegralform}
f(x)  = \int_{-1}^{+1} w(\mathcal H_a(\mathcal U), f, x, z) \intd z
+
\underbrace{a f(-1) k_a(x,-1)-a f(1)k_a(x,1)}_{\equiv B(\mathcal H_a(\mathcal U), f,x)}.
\end{align}
\end{lemma}
\begin{proof}
\prooflabel{\sref{Lemma}{lemma:compactintegralform:main}}
Integration by parts gives $\langle f, \partial g\rangle = \lsb fg \rsb_{-1}^{+1} - \langle \partial f, g\rangle $. Hence, the boundary terms involving derivatives of $\phi_x$ vanish and we have
\begin{align}
& \int_{-1}^{+1} w_{[-1,+1]}(\mathcal H_a, f, x, z) \intd z
\equiv \langle P \phi_x, P f \rangle  
= 
\langle (\mathbbm 1 + a \partial) \phi_x, (\mathbbm 1 + a \partial) g \rangle 
\\ & ~~
= 
\langle \phi_x, f \rangle 
+
a \langle \phi_x, \partial f \rangle 
+
a \langle \partial \phi_x,  f \rangle 
+
a^2
\langle \partial \phi_x, \partial f \rangle 
\\ & ~~ = 
\langle \phi_x, f \rangle 
+
a \big( \lsb \phi_x f\rsb_{-1}^{+1} - \langle \partial \phi_x, f \rangle \big)
+
a \langle \partial \phi_x,  f \rangle 
+
a^2
\big( \lsb \partial \phi_x f\rsb_{-1}^{+1} - \langle \partial^2 \phi_x, f \rangle \big)
\\ & ~~
= 
\langle \phi_x, f \rangle 
+
a \lsb \phi_x f\rsb_{-1}^{+1} 
-
a^2 \langle \partial \phi_x, \partial f \rangle
= 
\langle P^* P \phi_x, f \rangle 
+
a \lsb \phi_x f\rsb_{-1}^{+1},
= 
\langle \delta_x, f \rangle 
+
a \lsb \phi_x f\rsb_{-1}^{+1},
\end{align}
which rearranges to \eqref{eqn:compactintegralform}.
\end{proof}

The boundary condition terms are handled by the trivial
\begin{lemma}
	\label{lemma:bcsresidual}
	$M_n(F_{\mathcal H_a(\mathcal U)})$ is a pairwise stochastic gradient estimator for $x\mapsto f(x)-\big(B(\mathcal H_a, f,x)\big)$.
\end{lemma}
	Adding \textit{e.g.} the standard log-derivative trick estimator for $\nabla_\theta \expect{x\sim p(\cdot|\theta)}{B(\mathcal H_a, f,x)}$ to $M_n(F_{\mathcal H_a(\mathcal U)})$ therefore yields an estimator for $\nabla_\theta \expect{x\sim p(\cdot|\theta)}{f(x)}$. The properties of this estimator may be analysed analogously to \sref{Theorem}{thm:combination:mnln}, though this analysis provides little additional insight.

\section{Multivariate Extension}

\label{sec:multivariate}

We now give multivariate generalisations of the $w$ of \eqref{eqn:w:sym} and \eqref{eqn:w:bounded:closedform} which define our general family of pairwise representation functions \eqref{eqn:mn:f:fh}.

For $\realset^d$ we have, as proven in \sref{Appendix}{proof:multivariate:rn}, 

\begin{lemma}
\label{lem:multivariate:noncompact}
\theoremname{Integral representation on $\realset^d$}
For any $f:\realset^d\rightarrow \realset$ which is sufficiently differentiable and for which for any $\mathcal J \subseteq \lcb 1, 2, \dots, d\rcb, \quad \lim_{\norm{\bm z}_1 \to + \infty}  \abs{ \frac  {\partial ^{ \abs{\mathcal{J}}}f}{\prod_{k \in \mathcal J} \partial z_k}(\bm z)} < + \infty $, for any $\bm x \in \realset^d$ 
and $a>0$ we have, where $\mathrm{const}$ is constant with respect to $ \bm x$,
\begin{align}	
\label{eqn:multivariate:noncompact}
	& f( \bm x) = \\ & \int_{\realset^d} 
	\underbrace{
	\frac{ %
	\sum_{\substack{\mathcal J \subseteq \lcb 1, 2, \dots, d\rcb}} 
	a^{\abs{\mathcal{J}}} \frac  {\partial ^{\abs{\mathcal{J}}}f}{\prod_{k \in \mathcal J}\partial z_k}( \bm z)\prod_{k \in \mathcal J}\sign( x_k- z_k) %
	\exp(-\frac{\norm{ \bm x- \bm z}_1}{a})}{(2a)^{d}}
	}_{\equiv w(\mathcal H_a(\realset^d), f, \bm x, \bm z)}
	\intd \bm z + \mathrm{const.}
\end{align}
\end{lemma}

Similarly to \eqref{eqn:symmetriseft} the constant in the above representation does not affect our estimator. Although the above representation involves all partial derivatives up to order $d$, crucially this is the dimension of the sample space, rather than that of the parameter space. For example, in a variational auto-encoder \citep{vae,pmlr-v32-rezende14}, $d$ is the dimension of the low-dimensional embedding, which is often set to $d=2$ for visualisation purposes.

For the compact case (see the comment after \sref{Lemma}{lemma:mn:h:fh:representation} on the relevance) we have, as proven in \sref{Appendix}{proof:multivariate:compact},
\begin{lemma}
\theoremname{Integral representation on $[-1,1]^d$}
\label{lem:multivariate:compact}
Let $f:[-1,1]^d\rightarrow \realset$ be $d$-times differentiable and $a>0$. Along with $k_a$ of \eqref{eqn:agnank} define 
\begin{align}
	r_a(z,x)
	& \equiv
	\big(1+\sign(x-z)\tanh((1+\sign(x-z)z)/a)\big)
	\\
	\kappa_a(z,x)
	& \equiv 
	r_a(z,x)k_a(z,x)
\\
	  B_a(x) & =\cosh((1+x)/a)/\sinh(2/a)
	\\
	  \kappa_a(\bm z, \bm x) &  \equiv \prod_{i}^d \kappa_a(z_i,x_i).
\end{align}
Then %
we have the integral representation
\begin{align}	
\label{eqn:multivariate:compact}
	& f(\bm x) = \\
	& \int_{\mathbb [-1,1]^d} 
	\underbrace{ 
	\sum_{\mathcal J \subseteq \lcb 1, 2, \dots, d\rcb }
	\biggl(a^{\abs{\mathcal{J}}} \frac  {\partial ^{\abs{\mathcal{J}}}f}{\prod_{k \in \mathcal J}\partial z_k}(\bm z) \biggr) \kappa_a(\bm z, \bm x)
	}_{\equiv w(H_a([-1,+1]^d), f, \bm x, \bm z)} \intd  \bm z 
	\hspace{4mm}-\hspace{-6mm} \sum_{\bm \epsilon \in \{-1,0,1\}^d\setminus\lbrace{(0,\dots ,0)}\rbrace}
	B_a^{\epsilon}(\bm x) \cdot f\big(\bm x^{\epsilon} \big),
\intertext{where $\bm x^{\epsilon} = \begin{pmatrix} \epsilon_1 + (1-\epsilon_1)(1+\epsilon_1) x_1 \\ \cdots \\  \epsilon_d + (1-\epsilon_d)(1+\epsilon_d) x_d\end{pmatrix} $ and $B_a^{\epsilon}(\bm x) = \prod_{i=1}^d B_a(\epsilon_i x_i)\epsilon_i + (1-\epsilon_i)(1+\epsilon_i). $}
\end{align}
\end{lemma}

The terms involving $B_a^\epsilon(\cdot)$ may be handled analagously to the statement following \sref{Lemma}{lemma:bcsresidual}.

\section{Illustrative Example}
\label{sec:experiments}

We now demonstrate our new estimators both analytically and numerically on a simple univariate problem. The toy distribution $p(x|\theta)$ which we consider is designed with the following considerations:
\begin{itemize}
	\item The choice of distribution side-steps issues of high variance due to the factor of $\frac 1 {p(x|\theta)}$ in the fundamental trick representation \eqref{eqn:mn:f:ft} which may arise in the case of small values of $p(x|\theta)$. This is achieved by choosing a distribution that, on it's support, is bounded away from zero.
	\item The distribution is non-trivial to handle as it fails to satisfy the absolute continuity requirement of \autoref{lemma:unbiasedness}. We demonstrate how to decompose the distribution to make it amenable to both the log-derivative estimator and our new pairwise ones.
\end{itemize}
\label{sec:experiments:analytical}
\noindent \subparagraph{Toy Problem.}
Consider $f(x)=x$ where $x,\theta\in\mathbb{R}$. We wish to estimate \eqref{eqn:goal} at the point $\theta=0$, with
\begin{align}
p(x|\theta)
& = 
\frac{2}{\pi} \begin{cases}
1/(1+(x-\theta)^2) & \abs{x-\theta} \leq 1 \\
0 & \text{otherwise.}
\label{eqn:prational}
\end{cases}
\end{align}
\noindent \subparagraph{Reparameterisation Solution.} 
Letting $\epsilon$ of \eqref{eqn:reparamtrick:sum} have the probability density \eqref{eqn:prational} with $\theta=0$, \ie\
\begin{align}
p(\epsilon)
& = 
\frac{2}{\pi} \begin{cases}
1/(1+\epsilon^2) & \abs{\epsilon} \leq 1 \\
0 & \text{otherwise,}
\end{cases}
\end{align}
we may reparameterise
\begin{align}
    t_\theta(\epsilon) = x+\theta.
\end{align}
We then have by \eqref{eqn:reparamtrick:sum} that
\begin{align}
\mu\equiv \nabla_\theta \expect{\bm x \sim p(\cdot|\theta)}{f(\bm x)}
 = 
\expect{\bm \epsilon \sim \mathcal E}{\nabla_\theta f(t_\theta(\bm \epsilon))} 
=
\expect{\bm \epsilon \sim \mathcal E}{1} = 1. 
\label{eqn:muvaluereparam}
\end{align}

\noindent \subparagraph{Absolutely Continuous Decomposition.}
It is instructive to note that \eqref{eqn:prational} does not satisfy the absolute continuity requirement of \autoref{lemma:unbiasedness} (nor that of the log-derivative estimator \citep{glasserman2004monte,pflugbook,mohamed2019monte}), as the support depends on $\theta$. By introducing a simple decomposition, however, we can apply both our pairwise estimators and the log-derivative estimator. To this end we define (by plugging into \eqref{eqn:prational})
\begin{align}
c\equiv p(-1|\theta=0)=p(+1|\theta=0)=\frac 1 \pi,
\end{align}
and write the Heaviside function $u$ as
\begin{align}
    u(z)=
    \begin{cases}
        1 & \text{if} -1 \leq z \leq 1, \\
        0 & \text{otherwise.}
    \end{cases}
\end{align} 
Since $\theta$ translates the distribution, we have
\begin{align}
    p(x|\theta=\widehat\theta) & = p(x-\widehat\theta|\theta=0) u(x-\widehat\theta) \\
    & = (p(x-\widehat\theta|\theta=0) - c + c) u(x-\theta) \\
    & \equiv \hat p(x-\widehat\theta) + c u(x-\widehat\theta),
\end{align}
where by construction $\widehat p(x)\equiv p(x|\theta=0) - c$ satisfies the absolute continuity condition placed on $p$ in the statement of \autoref{lemma:unbiasedness}, and so we have
\begin{align}
    \mu & \equiv \nabla_\theta \mathbb E_{x\sim p(\cdot|\theta)}[f(x)] \\
    & = \nabla_\theta \int p(x|\theta) f(x) \intd x \\
    & = \nabla_\theta \int \big(\widehat p(x-\theta) + c u(x-\theta) \big)f(x) \intd x  \\
    & = \underbrace{\nabla_\theta \int \widehat p(x-\theta) f(x) \intd x}_{\equiv \widehat \mu} + \underbrace{\nabla_\theta \int c u(x-\theta) f(x)\intd x}_{=c(f(1)-f(-1))=\frac 2 \pi},
    \label{eqn:muhatdecomp}
\end{align}
where since $\nabla_\theta$ absorbs additive constants in $\widehat p$, we see that $\widehat \mu$ matches the expectation of the log-derivative family of estimators (including our pairwise ones) applied naively to $p(x|\theta)$.
\noindent \subparagraph{Log-derivative Trick.}
The remaining term is then available in closed form, as
\begin{align}
    \widehat \mu
    & = \int_{-1}^1 p(x|\theta=0) f(x) \nabla_{\theta=0} \log p(x|\theta)\intd x \\    
    \label{eqn:muhathalfway}
    & = \int_{-1}^1 \frac{4 x^2}{\pi (1+x^2)^2)} \intd x \\    
    & = 1-\frac 2 \pi. 
    \label{eqn:muhatvalue}
\end{align}
Hence by combining \eqref{eqn:muhatdecomp} with \eqref{eqn:muhatvalue} we have
\begin{align}
    \mu = \widehat \mu + \frac 2 \pi = 1-\frac 2 \pi + \frac 2 \pi = 1,
\end{align}
in line with \eqref{eqn:muvaluereparam}.
\noindent \subparagraph{Fundamental Trick.}
Consider also the fundamental trick estimator with the minimal two points, \ie\ $F_2$ of \autoref{lemma:unbiasedness}. Then it is straightforward to show that it reduces to the form \eqref{eqn:muhathalfway}, and therefore also gives the correct expectation.
\noindent \subparagraph{Further Numerical analysis.}
We provide a detailed numerical analysis of the variance of the main estimators under consideration in this paper, for the problem of estimating $\widehat \mu$ in \sref{Appendix}{sec:experiments:analytical:details}. We find that mixing an optimal $\approx 83\%$ of the log-derivative estimator with the remainder the fundamental trick estimator yields $\approx 7\%$ variance reduction for $n=2$ Monte Carlo samples, compared with the log-derivative estimator. While a modest improvement, \Autoref{fig:symdemo} shows greater gains  for larger $n$ due to averaging $O(n^2)$ terms in \eqref{eqn:logreparameterisationtrick:scalar:doublesum} --- moreover these gains do not require convex combination, but rather show our fundamental trick outperforming the log-derivative trick outright. %

\section{Conclusion}
\label{sec:conclusion}
We introduced a family of \textit{pairwise stochastic gradient estimators} for the problem of estimating gradients of an expectation with respect to the parameters of the distribution. This provides a fresh avenue of exploration for a fundamental sub-problem of various technical fields, notably that of Bayesian machine learning. We have presented a theoretical analysis of this new approach along with a set of numerical experiments which provide further intuitions and proof of concept. The work raises a number of important questions, however. Chief among these is how best to apply the new estimators %
on real-world problems for which the number of required samples is prohibitive, %
such as reinforcement learning of physical systems via policy gradient methods.

\bibliography{walder}
\bibliographystyle{apalike}

\clearpage

\appendix

The following are appendices for \textit{\thetitle}.

\section{A Constructive Derivation of the Fundamental Trick}
\label{sec:gore}

The setup for this approach is as follows.
Consider without loss of generality evaluating \eqref{eqn:goal} at $\theta=0$. With a slight abuse of notation (including reuse of $t_\theta$ from \eqref{eqn:reparamtrick:sum}), 
let the mapping $t_\theta(x)$ transform $x_0\sim p(\cdot|\theta=0)$ to $x = t_\theta(x_0)$ such that $x\sim p(x|\theta)$, \ie , by the transformation of random variables we have the relation involving the Jacobian determinant
\begin{align}
\label{eqn:firstorder:a}
\smash{
p(x|\theta)
 = 
p(t_\theta\mo(x)|\theta=0)
\abs{\frac{\partial}{\partial x} t_\theta\mo(x)}
}.
\end{align}

Expand $p(x|\theta)$ to first order with respect to $\theta$
\begin{align}
\label{eqn:series:theta}
\smash{
p(x|\theta) 
 =
p(x|\theta=0)
+
\theta \left.\frac{\partial}{\partial \theta}\right|_{\theta=0}p(x|\theta)
+
\mathcal O(\theta^2)},
\end{align}
and expand $t_\theta$ with respect to $\theta$ about $t_0(x)=x$, 
\begin{align}
\label{eqn:series:t}
t_\theta(x)
& =
x
+
\theta \Delta(x)
+\mathcal O(\theta^2)
.
\end{align}

By neglecting the second order terms, and applying Monte Carlo to the resultant expectation, we are able to obtain with some algebra (see \sref{Appendix}{sec:gore}), the estimator
\begin{align}
\label{eqn:logreparameterisationtrick:scalar:singlesum}
 \nabla_\theta \expect{x \sim p(\cdot|\theta)}{f(x)} \approx 
 \frac{1}{n}
\sum_{i=1}^n
\frac{f'(z_i)}{p(z_i|\theta)} \mathbb I (x_i>z_i) \frac{\partial}{\partial \theta}\log p(x_i|\theta),
\end{align}
where $x_i, z_i \sim p(\cdot|\theta)$. By symmetrising \eqref{eqn:logreparameterisationtrick:scalar:singlesum} with both \eqref{eqn:symmetrisepairs} (to introduce the double sum) and \eqref{eqn:symmetriseft} (to introduce the $\sign$ function), we obtain the main version of the fundamental trick estimator, \eqref{eqn:logreparameterisationtrick:scalar:doublesum}.

Expansions \eqref{eqn:series:theta} and \eqref{eqn:series:t} are coupled by \eqref{eqn:firstorder:a}; equating these two expansions of $p(x|\theta)$ and letting $u_\theta(x)\equiv t_\theta\mo(x)$ gives  
\begin{align}
p(x|\theta=0)
+
\theta \left.\frac{\partial}{\partial \theta}\right|_{\theta=0}p(x|\theta)
\label{eqn:firstorder:b}
 = 
 p(u_\theta(x)|\theta)
\abs{u_\theta'(x)} + \mathcal O(\theta^2).
\end{align}

By construction, \eqref{eqn:firstorder:b} holds at $\theta=0$; this can be verified by letting $\theta=0$ and observing that the Jacobian $\abs{u'_\theta(x)}$ is equal to 1 at $\theta=0$ (indeed, $u_0(x)=x$ is the identity).

Consider infinitesimal $\theta$ (\textit{i.e.} neglect $\mathcal O(\theta^2)$); differentiating w.r.t. $\theta$ at $\theta=0$ gives the additional condition
\begin{align}
p(x|\theta=0) & = p(x|\theta=0) \\
\left.\frac{\partial}{\partial \theta}\right|_{\theta=0}
\lnb
p(x|\theta=0)
+
\theta \left.\frac{\partial}{\partial \theta}\right|_{\theta=0} p(x|\theta)
+ \mathcal O(\theta^2)
\rnb
& = 
\left.\frac{\partial}{\partial \theta}\right|_{\theta=0}
\Big(
p(u_\theta(x)|\theta)
\abs{u_\theta'(x)} + \mathcal O(\theta^2)
\Big)
\end{align}
which simplifies to
\begin{align}
\label{eqn:firstorder:c}
\left.\frac{\partial}{\partial \theta}\right|_{\theta=0} p(x|\theta)
& = 
\left.\frac{\partial}{\partial \theta}\right|_{\theta=0}
p(u_\theta(x)|\theta)
\abs{u_\theta'(x)}.
\end{align}
For small $\theta$, the inverse $u_\theta(x)=x-\theta \Delta(x)$ for some function $\Delta(\cdot)$ (identified below), because
\begin{align}
\left.\frac{\partial }{\partial \theta}\right|_{\theta=0} t_\theta(u_\theta(x))
& =
\left.\frac{\partial }{\partial \theta}\right|_{\theta=0} \lnb
u_\theta(x)
+
\theta \Delta(u_\theta(x))
+\mathcal O(\theta^2)
\rnb
\\
& =
\left.\frac{\partial }{\partial \theta}\right|_{\theta=0} \lnb
\lnb x - \theta \Delta(x) \rnb
+
\theta \Delta(x - \theta \Delta(x) )
+\mathcal O(\theta^2) 
\rnb
\\
& = 0
\end{align}
as required. Hence, to handle the r.h.s. we exploit our choice of  \ie\ 
{
\small
\begin{align} 
& 
\left.\frac{\partial}{\partial \theta}\right|_{\theta=0}
p(u_\theta(x)|\theta)
 \abs{u_\theta'(x)}
\\ & = 
\left.\frac{\partial}{\partial \theta}\right|_{\theta=0}
\lnb
p(x - \theta \Delta(x)|\theta)
\abs{\frac{\partial}{\partial x}(x-\theta \Delta(x))}
\rnb
\\ & = 
\lnb 
\left.\frac{\partial}{\partial \theta}\right|_{\theta=0}
p(x - \theta \Delta(x)|\theta) 
\rnb
\lnb \abs{\frac{\partial}{\partial x}(x-0 \Delta(x))}\rnb
+ 
\Big( p(x - 0 \Delta(x)|\theta)\Big)
\left.\frac{\partial}{\partial \theta}\right|_{\theta=0}
\abs{\frac{\partial}{\partial x}(x-\theta \Delta(x))}
\\ & = 
\lnb  
\left.\frac{\partial (x-\theta \Delta(x))}{\partial \theta}\right|_{\theta=0}
\frac{\partial p(x - \theta \Delta(x)|\theta)}{\partial (x-\theta \Delta(x))}
 \rnb
\Big( 1 \Big)
-
p(x|\theta=0)
\frac{\partial}{\partial x} \Delta(x) \sign\lnb\lnb 1+0 \frac{\partial}{\partial x} \Delta(x)\rnb\rnb
\\ & ~~= 
- \Big(\frac{\partial}{\partial x}  p(x|\theta=0)\Big) \Delta(x)
-
 p(x|\theta=0) \Big( \frac{\partial}{\partial x}\Delta(x)\Big)
\\ & ~~= 
- \frac{\partial}{\partial x} \Big( p(x|\theta=0) \Delta(x) \Big).
\end{align}
}
Putting this back into the right hand side of \eqref{eqn:firstorder:c} gives
\begin{align}
\label{eqn:firstorder:d}
\left.\frac{\partial}{\partial \theta}\right|_{\theta=0} p(x|\theta)
& = 
- \frac{\partial}{\partial x} \Big( p(x|\theta=0) \Delta(x) \Big),
\end{align}
which is solved by integrating both sides w.r.t $x$, yielding
\begin{align}
\Delta(x) 
& = 
- \frac{1}{p(x|\theta=0)}
\int_{z=-\infty}^x 
\frac{\partial}{\partial \theta}p(z|\theta=0) 
\intd z
\\
& = 
+ \frac{1}{p(x|\theta=0)}
\int_{z=x}^\infty 
\frac{\partial}{\partial \theta}p(z|\theta=0) 
\intd z,
\end{align}
where we used 
\begin{align}
	\int_{z=-\infty}^x 
\frac{\partial}{\partial \theta}p(z|\theta=0) 
\intd z
+
\int_{z=x}^\infty 
\frac{\partial}{\partial \theta}p(z|\theta=0) 
\intd z
= 
\frac{\partial}{\partial \theta}
\int_{z=-\infty}^\infty 
p(z|\theta=0) 
\intd z
= 
\frac{\partial}{\partial \theta}
1
= 0.
\end{align}
We rewrite $\Delta$ as an expectation using the indicator 
$\mathbb I (x) \equiv 
\begin{cases}
1 & x \Leftrightarrow \text{True} \\
0 & \text{otherwise}
\end{cases}
$, to obtain%
\begin{align}
\Delta(x) 
& =
\frac{1}{p(x|\theta)}
\int_{-\infty}^{+\infty}
\mathbb I (z>x)
\frac{\partial}{\partial \theta}p(z|\theta=0) 
\intd z
\\
& =
\frac{1}{p(x|\theta)}
\frac{\partial}{\partial \theta}
\int_{-\infty}^{+\infty}
\mathbb I (z>x)
p(z|\theta=0) 
\intd z
\\
\label{eqn:delta:expect}
& = 
\frac{1}{p(x|\theta)} 
\mathbb E_{z\sim p(\cdot|\theta=0)}\hspace{-1mm}\lsb\frac{\partial}{\partial \theta}\log p(z|\theta=0)  \mathbb I (z>x)\rsb\hspace{-1mm}.
\end{align}
The chain rule with $t_\theta$ gives %
\begin{align}
\at{\frac{\partial}{\partial \theta}}{\theta=0} \mathbb E_{x\sim p(\cdot|\theta)}
\lsb
f(x)
\rsb
& = 
\mathbb E_{x\sim p(x|\theta=0)}
\lsb
\at{\frac{\partial}{\partial \theta} f(t_\theta (x))}{\theta=0}
\rsb
\\ & = 
\mathbb E_{x\sim p(x|\theta=0)}
\lsb
\at{\frac{\partial}{\partial \theta}}{\theta=0} f(x+\theta \Delta(x))
\rsb
\\ & = 
\mathbb E_{x\sim p(x|\theta=0)}
\lsb
\lnb \frac{\partial}{\partial x} f(x)\rnb \Delta(x)
\rsb
\\ & = 
\mathbb E_{x\sim p(x|\theta=0)}
\lsb
\frac{f'(x)}{p(x|\theta)} 
\mathbb E_{z\sim p(x|\theta=0)}
\lsb
\frac{\partial}{\partial \theta}\log p(z|\theta=0)  \mathbb I (z>x)
\rsb\rsb
\\ & = 
\label{eqn:logreparameterisationtrick:scalar:expectation}
\mathbb E_{x,z\sim p(x|\theta=0)}
\lsb
\frac{f'(x)}{p(x|\theta)} 
\frac{\partial}{\partial \theta}\log p(z|\theta=0)  \mathbb I (z>x)
\rsb
.
\end{align}
This scheme applies at any $\theta$ (not only $\theta=0$). Hence by Monte Carlo estimation of the expectation, we obtain the unsymmetrised form of our fundamental trick estimator, \eqref{eqn:logreparameterisationtrick:scalar:singlesum}.

\newcommand\thescalea{0.27}
\newcommand\thescaleb{0.7}
\newcommand\thescalec{0.98}
\begin{figure*}[ht]%
  \centering
  \subfigure[log-derivative trick]{
  \includegraphics[page=2,width=\thescalea\textwidth]{figs/cartoons-crop.pdf}
  }
  \\
  \subfigure[reparameterisation trick]{
    \includegraphics[page=1,width=\thescaleb\textwidth]{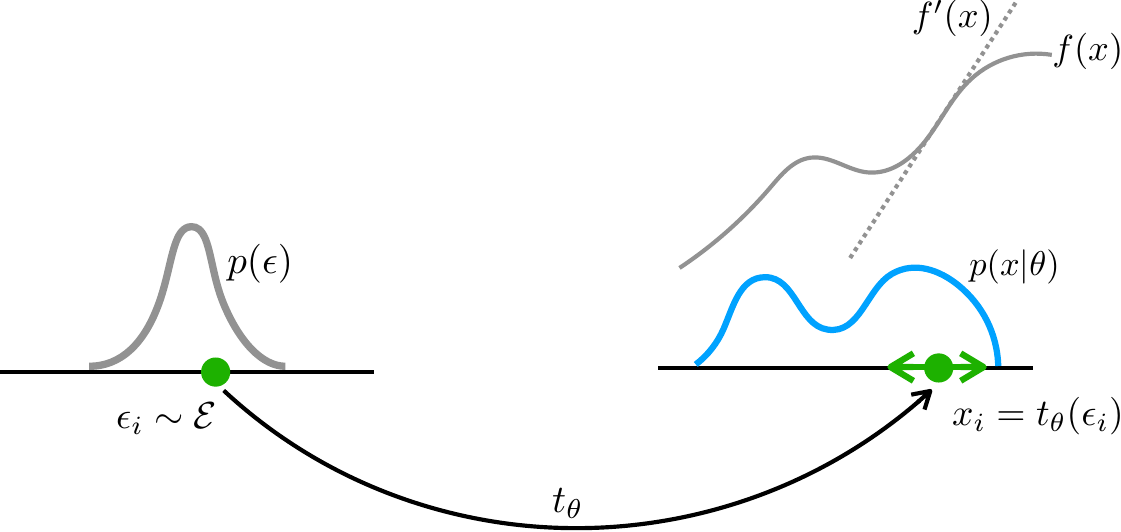}
  }
  \\
  \subfigure[fundamental trick]{
    \includegraphics[page=3,width=\thescalec\textwidth]{figs/cartoons-crop.pdf}
  }
  \caption{
    \textit{A sketch to accompany \sref{Appendix}{sec:gore}.
    } 
    \newline \textbf{(a)}~~ In the log-derivative trick, varying $\theta$  varies (as in the blue arrows) the probability density at the sampled point $x_i$ (denoted by the green dot), and thereby cannot make use of the derivative of $f$, but rather uses the value of $f(x_i)$ to weight $\nabla_\theta \log p(x_i|\theta)$ --- \textit{c.f.} \eqref{eqn:logtrick:sum}. 
    \newline \textbf{(b)}~~  In the reparameterisation trick, a sample $\epsilon_i$ from a fixed base distribution $\mathcal E$ is mapped by a deterministic function parameterised by $\theta$. Hence, varying $\theta$ perturbs the mapped point (as in the green arrows). The derivative of $f$ indicates the direction in which to perturb $x_i$ (and by the chain rule, $\theta$) in order to increase the value of $f(x_i)$ --- \textit{c.f.} \eqref{eqn:reparamtrick:sum}. 
    \newline \textbf{(c)}~~  In our derivation of the fundamental trick of \sref{Appendix}{sec:gore}, we initially sample an $x_i^{(0)}$ given some fixed value of the parameters (here and without loss of generality, $\theta=0$). Unlike the reparameterisation trick, we have no explicit transformation $t_\theta$, but by considering infinitesimal changes to $\theta$ (the r.h.s. red and blue plots denote $p(x|\theta=0)$ and $p(x|\theta)$, respectively) we construct an appropriate infinitesimal perturbative transformation of the $x_i^{(0)}$, namely \eqref{eqn:series:t}. Hence changing $\theta$ may be thought not only to perturb the mapped point (as denoted by the green arrows), but also to increases or decreases the probability of the mapped point (as denoted by the blue arrows). In this way, both the derivative of $f$ and the gradient of the (log of the) density $p(x|\theta)$ are made use of --- \textit{c.f.} \eqref{eqn:logreparameterisationtrick:scalar:singlesum}.
    \label{fig:cartoon}
  }
\end{figure*}
\FloatBarrier

\section{Hilbert-Sobolev-Laplace Gradient Estimator}
\label{sec:hsl}

To contrast with the novel and theoretically useful estimators of the main paper, we present here a tempting alternative which we show to be essentially useless, in the sense that it does not offer any orthogonal component of information over and above the log-derivative estimator.

\subsection{The Estimator}
\label{sec:hsl:definition}
To eliminate the factor $\frac{1}{p(z|\theta)}$ in \eqref{def:mn:f:rkhs} it is tempting to employ a pairwise estimator using pairs $(x, z)$, where we let $x\sim p(\cdot|\theta)$ and, instead of $z\sim p(\cdot|\theta)$ as per \eqref{eqn:pairwisevariance:mn},
\begin{align}
p(z|x) 
 = 
\laplace (z|x,a)
\label{eqn:hslproposal}
 \equiv
\exp(-\abs{x-z}/a)/(2a).
\end{align}
The point is that the factor then cancels with part of the weight function $w$ of \eqref{eqn:w:sym}, and it is easy to show using a similar argument as the proof of \sref{Lemma}{thm:pairwiseunbiasedness} that the following \textit{Hilbert-Sobolev-Laplace estimator}
\begin{empheq}[box=\ovalbox]{align}
\label{eqn:hsl:univariate}
\nabla_\theta \expect{x \sim p(\cdot|\theta)}{f(x)}
& \approx
\frac{1}{mn}
\sum_{i=1}^m
\sum_{j=1}^n 
\lnb f(x_{i,j}) + a \sign (x_i-x_{i,j}) f'(x_{i,j}) \rnb
\nabla_\theta \log p(x_i|\theta),
\end{empheq}
is unbiased, where $x_i \sim p(\cdot|\theta)$ as before, but now $x_{i,j} \sim \laplace(x_i,a)$ has mean $x_i$ and scale $a$. 

\subsection{Uselessness of the Estimator}

Does \eqref{eqn:hsl:univariate} offer any independent information over the log-derivative estimator \eqref{eqn:logtrick:sum}? No; in line with \sref{Lemma}{lemma:momentcondition} that would require $\hat\Sigma_\mathrm{hl}\neq\hat\Sigma_\mathrm{ll}$, for the (uncentered) second moments (indexed by l and h for the log-derivative and Hilbert-Sobolev-Laplace estimators, respectively), but
\begin{align}
\hat\Sigma_\mathrm{ll}
& = 
\expectlr{x\sim p(\cdot|\theta)}{\big( f(x)\nabla \log p(x|\theta) \big)^2}
\end{align}
and
\begin{align}
\hat\Sigma_\mathrm{hl}
& = 
\expectlr{
x\sim p(\cdot|\theta)
}{
\expectlr{z\sim \laplace(x,a)}{
\big(f(z)+a\sign(x-z)f'(z)\big) \nabla \log p(x|\theta) f(x)\nabla \log p(x|\theta) }}
\\
& = 
\expect{
x\sim p(\cdot|\theta)
}{
f(x)\big( \nabla \log p(x|\theta) \big)^2\,\,
\smash{\underbrace{\expect{z\sim \laplace(x,a)
}{
\big(f(z)+a\sign(x-z)f'(z) \big)
}
}_{= f(x)}}
}
\\
& = 
\hat\Sigma_\mathrm{ll},
\end{align}
where the underbrace is expounded upon in the following \sref{Appendix}{sec:laplaceidentity}. Note that we needed only consider the diversification between the log-derivative estimator associated with (in the notation of the previous \sref{Appendix}{sec:hsl:definition}) a single $x_i$ and the associated Hilbert-Sobolev-Laplace term associated with a single perturbation $x_{i,j}$ of that point.
 
 Hence while it is tempting to choose \eqref{eqn:hslproposal} in order to obtain the convenient looking \eqref{eqn:hsl:univariate}, this estimator is essentially just a noisy version the log-derivative estimator, and we cannot reduce the variance by taking a convex combination of the two. 
\subsection{Laplace Evaluation Estimator}
\label{sec:laplaceidentity}
As an aside, we note that putting the $w$ of \eqref{eqn:w:sym} into \eqref{eqn:reproducing} and writing the result as an expectation leads to the neat identity
\begin{align}
\label{eqn:laplaceidentity}
f(x) 
 = 
\expect{z\sim \laplace(x,a)}{f(z)+a f'(z)\sign(x-z)},
\end{align}
where $f'$ is the derivative, $\sign$ is the $\pm 1$ valued sign function, and the Laplace distribution has mean $x$ and scale $a$, that is $p(z|x,a)=\exp(-\abs{x-z}/a)/(2a)$.

\clearpage

\section{Lemmas and Proofs for \texorpdfstring{\sref{Section}{sec:theoretical}}{}}
\label{sec:varproofs}

We begin the law of total variance, that is
\begin{lemma}\theoremname{Eve's law.}
	\label{lemma:eve}
	Let $x$ and $y$ be random variables on the same probability space, and let $y$ have finite variance. Then we have the well known identity
	\begin{align}
		\label{eqn:eve}
		\varwrt{y}{y}=\expect{x}{\varwrt{y}{y|x}}+\varwrt{x}{\expect{y}{y|x}}.
	\end{align}
\end{lemma}
\begin{proof}
\prooflabel{\sref{Lemma}{lemma:eve}}
\begin{align}
 	\varwrt{y}{y} 
 	& =
 	\expect{y}{y^2}-\expect{y}{y}^2
 	\\
 	& =
 	\expect{x}{\expect{y}{y^2|x}}-\expect{x}{\expect{y}{y|x}}^2
 	\\
 	& =
 	\expect{x}{\varwrt{y}{y|x}+\expect{y}{y^2|x}}-\expect{x}{\expect{y}{y|x}}^2
 	\\
 	& =
 	\expect{x}{\varwrt{y}{y|x}}+\expect{x}{\expect{y}{y^2|x}}-\expect{x}{\expect{y}{y|x}}^2
 	\\
 	& = \expect{x}{\var{y|x}}+\varwrt{x}{\expect{y}{y|x}}.
 \end{align}
\end{proof}
\begin{corollary}
	\theoremname{Conditional Eve's law.}
	\label{cor:eve:conditional}
	By conditioning all distributions on a random variable $z$ on the same probability space as $x$ and $y$, we obtain
	\begin{align}
		\varwrt{y}{y|z}&=\expect{x}{\varwrt{y}{y|x,z}|z}+\varwrt{x}{\expect{y}{y|x,z}|z},
	\intertext{or equivalently}
	\label{eqn:eve:conditional}
		\varwrt{y}{y|x}&=\expect{z}{\varwrt{y}{y|z,x}|x}+\varwrt{z}{\expect{y}{y|z,x}|x}.
	\end{align}
\end{corollary}
\begin{corollary}
	\theoremname{Bivariate Eve's law.}
	\label{cor:eve:bivariate}
	Substituting \eqref{eqn:eve:conditional} in the first term on the r.h.s. of \eqref{eqn:eve} yields
	\begin{align}
		\varwrt{y}{y}
		& =
		\expectlr{x}{\expect{z}{\varwrt{y}{y|z,x}|x}+\varwrt{z}{\expect{y}{y|z,x}|x}}+\varwrt{x}{\expect{y}{y|x}}
		\\
		\label{eqn:eve:bivariate}
		& =
		\expectlr{x,z}{\varwrt{y}{y|z,x}|x}+\expectlr{x}{\varwrt{z}{\expect{y}{y|z,x}|x}}+\varwrt{x}{\expect{y}{y|x}}.~~~~
	\end{align}
\end{corollary}
\begin{lemma}
	\theoremname{Multivariate Eve's law.}
	\label{lemma:eve:multivariate}
	Let $x_1, x_2, \dots , x_n$ and $y$ be random variables on the same probability space, and let $y$ have finite variance. Then
	\begin{align}
		\varwrt{y}{y}
		& = 
		\expectlr{\xoneton}{\varwrt{y}{y|\xoneton}|\xonetonmo}
		\\
		& ~~~~ +
		\sum_{i=1}^{n}\expectlr{\xonetoimo}{\varwrt{x_i}{\expectlr{y}{y|\xonetoi}\xonetoimo}|\xonetoimtwo}
		\\
		& ~~~~ +
		\varwrt{x_1}{\expect{y}{y|x_1}}.
		\label{eqn:eve:multivariate}
	\end{align}
\end{lemma}
\begin{proof}
	\prooflabel{\sref{Lemma}{lemma:eve:multivariate}}
	Applying \sref{Corollary}{cor:eve:conditional} once again yields
	\begin{align}
	\label{eqn:eve:conditionaltwice}
		\varwrt{y}{y|x_1,x_2}
		& = 
		\expectlr{x_3}{\varwrt{y}{y|x_1,x_2,x_3}|x_1,x_2}
		+
		\varwrt{x_3}{\expectlr{y}{y|x_1, x_2, x_3}|x_1, x_2}. ~~~~~~
	\end{align}
	Relabelling terms in \eqref{eqn:eve:bivariate} we obtain
	\begin{align}
		\varwrt{y}{y}
		& =
		\expectlr{x_1, x_2}{\varwrt{y}{y|x_1,x_2}|x_1}+\expectlr{x_1}{\varwrt{x_2}{\expect{y}{y|x_1,x_2}|x_1}}+\varwrt{x_1}{\expect{y}{y|x_1}}.		
	\end{align}
	Substituting \eqref{eqn:eve:conditionaltwice} into the first term above, we obtain
	\begin{align}
		\varwrt{y}{y}
		& =
		\expectlr{x_1, x_2,x_3}{\varwrt{y}{y|x_1,x_2,x_3}|x_1,x_2}
		\\
		& ~~~~ +
		\expectlr{x_1, x_2}{\varwrt{x_3}{\expectlr{y}{y|x_1, x_2, x_3}|x_1, x_2}
		|x_1} 
		\\ 
		& ~~~~ +
		\expectlr{x_1}{\varwrt{x_2}{\expect{y}{y|x_1,x_2}|x_1}}
		\\ 
		& ~~~~ +
		\varwrt{x_1}{\expect{y}{y|x_1}}.
	\end{align}
	Repeating the process for $x_4, x_5, \dots, x_n$ yields the desired result.
\end{proof}
\begin{lemma}
	\label{lemma:eve:functional}
	\theoremname{Functional Eve's law.}
	Let $\xoneton$ be i.i.d. random variables with domain $\mathcal X$, and let $f:\mathcal X^n\rightarrow \realset$ be a deterministic function with $\varwrt{\xoneton}{f(\xoneton)} < \infty$. Then
	\begin{align}
		\varwrt{\xoneton}{f(\xoneton)}
		& = \sum_{i=2}^n \expectlr{\xonetoimo}{\varwrt{x_i}{\expectlr{\xipoton}{f(x_1, x_2, \dots, x_n)}}}
		\\ & ~~~~~ + \varwrt{x_1}{\expect{\xtwoton}{f(\xoneton)}}.
		\label{eqn:eve:functional}
	\end{align}
\end{lemma}
\begin{proof}
	\prooflabel{\sref{Lemma}{lemma:eve:functional}}
	Putting $y=f(\xoneton)$ in \sref{Lemma}{lemma:eve:multivariate}, we observe that the first term in \eqref{eqn:eve:multivariate} vanishes since $f$ is deterministic and therefore has zero variance when its argument is fixed.
\end{proof}
\begin{lemma}
	\theoremname{Inversion of expectation and variance.}
	\label{lem:inverting}
	Let $x, y$ and $z$ be random variables. Then
	\begin{align}
		\expectlr{x}{\varwrtlr{z}{y}} 
		= 
		\expectlr{z}{\varwrtlr{x}{ y - \expect{z}{y} }} 
		+\varwrt{z}{\expect{x}{y}}.
	\end{align}
\end{lemma}
\begin{proof}
	\prooflabel{\sref{Lemma}{lem:inverting}}
	Let $x, y$ and $z$ be random variables on the same probability space. Then we have the identity
	\begin{align}
		\expectlr{x}{\varwrtlr{z}{y}} 
		& =
		\expectlr{x}{\expectlr{z}{\lnb y-\expect{z}{y}\rnb^2}} 
		\\
		& =
		\expectlr{z}{\expectlr{x}{\lnb y-\expect{z}{y}\rnb^2}} 
		\\
		& =
		\expectlr{z}{\expectlr{x}{\lnb y-\expect{z}{y}\rnb^2}
		-
		\expectlr{x}{y-\expect{z}{y}}^2
		+
		\expectlr{x}{y-\expect{z}{y}}^2
		}
		\\
		& =
		\expectlr{z}{\expectlr{x}{\lnb y-\expect{z}{y}\rnb^2}
		-
		\expectlr{x}{y-\expect{z}{y}}^2
		+
		\lnb \expectlr{x}{y}-\expect{x,z}{y}\rnb^2
		}
		\\
		& = 
		\expectlr{z}{\varwrtlr{x}{y - \expect{z}{y}}} 
		+
		\expectlr{z}{\lnb\expectlr{x}{y}-\expect{x,z}{y}\rnb^2}
		\\
		& = 
		\expectlr{z}{\varwrtlr{x}{ y - \expect{z}{y}}} 
		+\varwrt{z}{\expect{x}{y}}.
	\end{align}
\end{proof}
\begin{proof}
	\prooflabel{\sref{Lemma}{lemma:eve:functional:inverted}}
	We apply the inversion of expectation and variance \sref{Lemma}{lem:inverting} with $x=(\xonetoimo), z=x_i$ and $y=\expectlr{\xipoton}{f(\xoneton)}$ to the summand of \eqref{eqn:eve:functional} to obtain
	\begin{align}
	& \expectlr{\xonetoimo}{\varwrt{x_i}{\expectlr{\xipoton}{f(\xoneton)}}}
	\\ & = \expectlr{x_i}{\varwrtlr{\xonetoimo}{\expectlr{\xipoton}{f(\xoneton)}-\expectlr{\xiton}{f(\xoneton)}}}
	\\ & ~~~~ + \varwrt{x_i}{\expectlr{\xmi}{f(\xoneton)}}.
	\end{align}
	Putting this into \eqref{eqn:eve:functional} yields the desired result.
\end{proof}
We now return our attention to the first of our main theoretical results, starting with the following
\begin{proof}
	\prooflabel{\sref{Theorem}{thm:pairwiseunbiasedness}}
	$M_n(G)$ is an unbiased estimator of $\nabla_\theta \expect{x \sim p(\cdot|\theta)}{f(x)}$ since 
	\begin{align}
		& \expectlr{}{M_n(G)}
		\stackrel{(a)}{=} 
		\expectlr{}{M_2} 
		\stackrel{(b)}{=} 
		\expectlr{x,z}{
		\half(
		G(x,z)\nabla_\theta \log p(x|\theta)
		+
		G(z,x)\nabla_\theta \log p(z|\theta)
		)}
		\\ & ~~ \stackrel{(c)}{=} 
		\expectlr{x,z}{
		G(x,z)\nabla_\theta \log p(x|\theta)}
		\stackrel{(d)}{=} 
		\expectlr{x}{
		f(x) \nabla_\theta \log p(x|\theta)
		}
		\stackrel{(e)}{=} 
		\nabla_\theta \expectlr{x}{
		f(x)}.
	\end{align}
	Here, ($a$) follows by linearity of expectation and summation, ($b$) is by definition, ($c$) is due to the observation that, for any function $g$ and i.i.d. random variables $x$ and $z$, $\expectlr{x,z\sim \mathcal D}{g(x,z)} = \expectlr{x,z\sim \mathcal D}{g(z,x)}$, ($d$) follows by definition of $G(\cdot, \cdot)$, and ($e$) is the standard log-derivative trick.
\end{proof}
\begin{proof}
	\prooflabel{\sref{Theorem}{thm:combination:mnln}}.
	The proof is similar to that of \ref{thm:pairwisevariance}, but with $M_n(G)$ replaced by $\alpha M_n(G) + (1-\alpha)L_n$ in the application of \sref{Lemma}{lemma:eve:functional:inverted}.
	First consider \eqref{eqn:varsummand}: in step $(c)$, we should include terms with $i=j$, which arise from the $L_n$. However, in step $(e)$ we remove those terms with $j\geq i$ as they are constant under the variance w.r.t. $\xonetoimo$. This removes the terms arising from $L_n$. We are left with the factor $\alpha$, which we move outside the variance to obtain the $\mathcal O(\frac{\alpha^2}{n^2})$ term in \eqref{eqn:combination:mnln:variance}.
	Now consider \eqref{eqn:varsummand:diag}. In step $(a)$ we once again include the diagonal terms, which in this case are those with $j=k$, which do not vanish in subsequent steps as explained in relation to \eqref{eqn:varsummand}. We therefore have
		\begin{align}
		 & \varwrtlr{x_i}{\expectlr{-i}{\alpha M_n + (1-\alpha)L_n}}
		 \\ & \stackrel{(a)}{=}
		 \varwrt{x}{\expect{z}{
		 \frac{1}{n}
		\lnb \alpha \lnb
		G(x,z)\nabla_\theta \log p(x|\theta)
		+
		G(z,x)\nabla_\theta \log p(z|\theta)
		\rnb + (1-\alpha) f(x) \nabla_\theta \log p(x|\theta) 
		\rnb
		}}
		 \\ & \stackrel{(b)}{=}
		\varwrt{x}{
		\frac{1}{n}
		\lnb \alpha \lnb
		f(x) \nabla_\theta \log p(x|\theta)
		+
		\expect{z}{G(z,x)\nabla_\theta \log p(z|\theta)
		\rnb + (1-\alpha) f(x) \nabla_\theta \log p(x|\theta) \rnb
		}}
		 \\ & \stackrel{(c)}{=}
		\frac{1}{n^2}
		\varwrt{x}{
		f(x) \nabla_\theta \log p(x|\theta)
		+
		\alpha \expect{z}{G(z,x)\nabla_\theta \log p(z|\theta)
		}},	
		\end{align}
		where $(a)$ follows from the previous argument with substitution of the definition of $M_n(G)$ and $L_n$, $(b)$ is by the definition of $G(x,z)$, and $(e)$ moves a constant outside the variance. Hence the summation on line \eqref{eqn:eve:functional:inverted:b} reduces to the first term of \eqref{eqn:combination:mnln:variance}, completing the proof.
\end{proof}
\begin{proof}
	\prooflabel{\sref{Theorem}{thm:varreduction}}
	We have the quadratic function of $\alpha$
	\begin{align}
	 	\varwrt{x\sim p(\cdot|\theta)}{f(x)\nabla_\theta \log p(x|\theta)+\alpha \expectlr{z\sim p(\cdot|\theta)}{G(z,x)\nabla_\theta \log p(z|\theta)}}
	 	= \sigma_L^2 + \alpha^2\sigma_F^2 + 2\alpha\rho\sigma_L \sigma_F,
	 \end{align} 
	where we have additionally defined 
	\begin{align}
		\sigma_F^2 & = \varwrt{x\sim p(\cdot|\theta)}{\expect{z\sim p(\cdot|\theta)}{G(x,z)\nabla_\theta \log p(z_\theta)}}.
	\end{align}
	Putting the optimal $\alpha^\star=-\rho \frac{\sigma_L}{\sigma_F}$ back into the quadratic and dividing by $\sigma_L$ yields the result.
\end{proof}

\clearpage

\section{Additional Mathematical Details for the Representer Tricks}

\subsection{First Principles Verification \texorpdfstring{of \eqref{eqn:w:sym}}{}}
\label{sec:firstprincipleslaplace}
We now check \eqref{eqn:w:sym} by putting it into \eqref{eqn:reproducing}
 and using integration by parts. This verifies both the symmetrisation and the original \rkhs\ based derivation. Assume w.l.o.g. that $x=0$, giving
\begin{align}
f(0)
 = 
\frac{1}{2a} \int_\mathbb R \exp\big(-\frac{|z|}{a}\big) (f(z)+a \sign (-z) f'(z)) \intd z
 = 
\frac{1}{2a} (I_1 + I_2),
\label{eqn:eyes}
\end{align}
where
$I_1 \equiv \int_{-\infty}^0 \exp(+z/a) (f(z)+af'(z)) \intd z$ and $I_2 \equiv \int_{0}^\infty \exp(-z/a) (f(z)-af'(z)) \intd z$. Then
\begin{align}
I_1 
& = 
\int_{-\infty}^0 \exp(\frac z a) (f(z)+af'(z)) \intd z
\\
& = 
\int_{-\infty}^0 \exp(\frac z a) f(z) \intd z + a \int_{-\infty}^0 \exp(+ \frac z a) f'(z) \intd z
\\
& = 
\int_{-\infty}^0 \exp(\frac z a) f(z) \intd z 
 + a \lnb \lsb \exp(\frac z a) f(z) \rsb_{-\infty}^{0} \hspace{-1mm} - \frac{1}{a} \int_{-\infty}^0 \exp(\frac z a) f(z) \intd z\rnb
\\
& = 
a \lsb \exp(+z/a) f(z) \rsb_{-\infty}^{0}
\\
& = a f(0),
\end{align}
and by a similar argument $I_2=a f(0)$, as required by \eqref{eqn:eyes}.

\subsection{Multivariate Integral Representations}

\label{proof:multivariate}

\subsubsection{Real Space \texorpdfstring{$\realset^d$}{}}
\label{proof:multivariate:rn}
For \sref{Lemma}{lem:multivariate:noncompact} we have the following

\begin{proof}
Let $f:[-1,1]^d\rightarrow \realset$ be $d$-times differentiable. Let define for any $\bm x$ in $[-1,1]^d$ and $a>0$: 
\begin{align}	
    \quad I(a,\bm x) \equiv \int_{\realset^d} 
	 \frac  {\partial ^{d}g_a}{\prod_{k \in \lcb 1, 2, \dots, d\rcb}\partial z_k}(\bm z)\prod_{k \in \lcb 1, 2, \dots, d\rcb}\sign( x_k - z_k) \intd \bm z,
\end{align}
where $g_a(\bm z) \equiv f(\bm z) \cdot \exp(-\abs{\bm x -\bm z}_1/a)$ for any $\bm z \in \realset^d \setminus\lbrace{\bm x}\rbrace$ and $ a>0$. 

$I$ is well defined because  $g_a$ is sufficiently differentiable on $\realset^d \setminus\lbrace{\bm x}\rbrace$.

The representation of $f$ follows from the above expression after having calculated it in two different ways. We will differentiate the expression within the integral and then we will integrate it. 

\newcommand\sss{\hspace{1.4mm}}
\newcommand\numeq[1]%
  {\stackrel{\scriptscriptstyle(\mkern-1.5mu#1\mkern-1.5mu)}{=}}
\subparagraph{Step 1: differentiation}

We differentiate $g$ with respect to each $z_i$ for $i \in \lcb 1, 2, \dots, d\rcb$. 
Set $ \bm z \in \realset^d \setminus\lbrace{\bm x}\rbrace$:
\begin{align}
	& \frac  {\partial ^{d}g_a}{\prod_{k \in \lcb 1, 2, \dots, d\rcb}\partial z_k}(\bm z)\prod_{k \in \lcb 1, 2, \dots, d\rcb}\sign(x_k - z_k)  \\ & \sss = \frac  {\partial ^{d}f(\bm z)\cdot \exp(-\frac{\norm{\bm x-\bm z}_1}{a})}{\prod_{k \in \lcb 1, 2, \dots, d\rcb}\partial z_k} \prod_{k \in \lcb 1, 2, \dots, d\rcb}\sign(x_k - z_k)
    \\ & \sss \numeq{1} \Bigg( \sum_{\substack{\mathcal J \subseteq \lcb 1, 2, \dots, d\rcb }} \frac  {\partial ^{\abs{\mathcal{J}}}f}{\prod_{k \in \mathcal J}\partial z_k} (\bm z) \frac  {\partial ^{d - \abs{\mathcal{J}}}\exp(-\frac{\norm{\bm x-\bm z}_1}{a})}{\prod_{k \notin \mathcal J}\partial z_k} \Bigg) \prod_{k \in \lcb 1, 2, \dots, d\rcb}\sign(x_k - z_k)
    \\ & \sss \numeq{2}  \Bigg( \sum_{\substack{\mathcal J \subseteq \lcb 1, 2, \dots, d\rcb \\ }} \frac  {\partial ^{\abs{\mathcal{J}}}f}{\prod_{k \in \mathcal J}\partial z_k} (\bm z) \frac{\prod_{k \notin \mathcal J}\sign( x_k- z_k)}{a^{(d - \abs{\mathcal{J}})}}\Bigg)  \exp(-\frac{\norm{\bm x-\bm z}_1}{a})  \prod_{k \in \lcb 1, 2, \dots, d\rcb}\sign(x_k - z_k)
    \\ & \sss = \frac{\bigg(\sum_{\substack{\mathcal J \subseteq \lcb 1, 2, \dots, d\rcb \\ }}  a^{\abs{\mathcal{J}}} \frac  {\partial ^{\abs{\mathcal{J}}}f} {\prod_{k \in \mathcal J}\partial z_k} (\bm z)\prod_{k \in \mathcal J}\sign( x_k-z_k) \bigg)\exp(-\frac{\norm{\bm x-\bm z}_1}{a})}{a^{d}}.
\end{align}

$(1)$ is given by the product rule.

$(2)$ is given by the the following formula, true for any  $ \bm z \in \realset^d \setminus\lbrace{\bm x}\rbrace$ and $i \in \lcb 1, 2, \dots, d\rcb$, $ \frac{\partial \exp(-\frac{\norm{\bm x-\bm z}_1}{a})}{\partial z_i} = \frac{\sign(\bm x-\bm z)}{a} \cdot\exp(-\frac{\norm{\bm x-\bm z}_1}{a})$ which we have applied $(d - \abs{\mathcal{J}})$-times.

So if we integrate the expression above with respect to $\bm z$ in $[-1,1]^d$,
\begin{align}
    & I(a,\bm x) = \int_{\realset^d} \Bigg( \frac{ \bigg(
    \sum_{\substack{\mathcal J \subseteq \lcb 1, 2, \dots, d\rcb }} 
    \big(a^{\abs{\mathcal{J}}} \frac  {\partial ^{\abs{\mathcal{J}}}f}{\prod_{k \in \mathcal J}\partial z_k}(\bm z)\prod_{k \in \mathcal J}\sign( x_k- z_k) \bigg) \exp(-\frac{\norm{\bm x-\bm z}_1}{a})}{a^d} \Bigg)\intd \bm z. \label{derivation_rn}
\end{align}
\subparagraph{Step 2: integration}
We integrate successively the derivatives inside the integral in $I$ with respect to each $z_i$ using the expression $\forall t \in \realset$,
\begin{align}
h(t) = \int_{\realset}h'(z)\frac{1}{2} \sign(t-z)\intd z + \frac{  \lim_{z \to +\infty}h (z) +  \lim_{z \to -\infty}h (z)  }{2}.     
\end{align}
Let's prove the integration for one integral and then let's repeat the integration successively by recurrence. For i in $\lcb 1, 2, \dots, d\rcb$, $\bm x$ in $ [-1,1]^{d}$ and $(z_2,\dots, z_d)$ in  $ [-1,1]^{d-1}$ then:
\begin{align}
    & \int_{\realset}  
    \frac  {\partial ^{d}g_a}{\prod_{k \in \lcb 1, 2, \dots, d\rcb}\partial z_k}(\bm z) \sign( x_1 - z_1) \intd z_1 = \\
     & ~ 2 \Bigg( \frac  {\partial ^{d-1}g_a}{\prod_{k \in \lcb 2, \dots, d\rcb}\partial z_k} ( x_1, z_2,\dots, z_d) 
    + \hspace{-1.5mm} \lim_{z_1 \to +\infty}  \frac  {\partial ^{d-1}g_a}{\prod_{k \in \lcb 2, \dots, d\rcb}\partial z_k} (\bm z) 
    +  \hspace{-1.5mm} \lim_{z_1 \to -\infty}  \frac  {\partial ^{d-1}g_a}{\prod_{k \in \lcb 2, \dots, d\rcb}\partial z_k} (\bm z)\Bigg).
\end{align}
We will show that under the conditions of the lemma, the last two terms are equals to 0. Under weaker conditions, the two last terms could be equal to a constant that depends on $a$ and $(z_2,\dots,z_d)$. According to the last step, we have the formula:
\begin{align}
    & \frac  {\partial ^{d-1}g_a}{\prod_{k \in \lcb 2, \dots, d\rcb}\partial z_k}  (\bm z)= \\
    & 
    ~~ \frac{\bigg(\sum_{\substack{\mathcal J \subseteq \lcb 2, \dots, d\rcb \\ }}  a^{\abs{\mathcal{J}}} \frac  {\partial ^{\abs{\mathcal{J}}}f}{\prod_{k \in \mathcal J}\partial z_k} (\bm z)\prod_{k \notin \mathcal J}\sign( x_k- z_k) \bigg) \exp(-\frac{\norm{\bm x-\bm z}_1}{a})}{a^{d-1}}.
\end{align}
As $\sum_{\substack{\mathcal J \subseteq \lcb 1, 2, \dots, (d-1)\rcb \\ }}  a^{\abs{\mathcal{J}}} \frac  {\partial ^{\abs{\mathcal{J}}}f}{\prod_{k \in \mathcal J}\partial z_k} (\bm z)\prod_{k \notin \mathcal J}\sign( x_k- z_k)$ is bounded for $\norm{\bm x} \to + \infty$, and 
\begin{align}
    \lim_{z_1 \to \pm \infty} \frac  {\partial ^{d-1}g_a}{\prod_{k \in \lcb 2, \dots, d\rcb}\partial z_k}  (\bm z) = 0,    
\end{align}
we get the result for one integral:
\begin{align}
    \int_{\realset}  
 \frac  {\partial ^{d}g_a}{\prod_{k \in \lcb 1, 2, \dots, d\rcb}\partial z_k} (\bm z) \sign( x_1 - z_1) \intd  z_1 
     = 2 \cdot \frac  {\partial ^{d-1}g_a}{\prod_{k \in \lcb 2, \dots, d\rcb}\partial z_k}(x_1,z_2,\dots,z_d).
\end{align}
Then we proceed by recurrence for $I(a,\bm x)$:
\begin{align}
    I(a,\bm x)  &  = \int_{\realset^d} 
	 \frac  {\partial ^{d}g_a}{\prod_{k \in \lcb 1, 2, \dots, d\rcb}\partial z_k}(\bm z)\prod_{k \in \lcb 1, 2, \dots, d\rcb}\sign( x_k - z_k) \intd \bm z
	\\ &  = \int_{\realset^{d-1}} \int_{\realset}  \frac  {\partial ^{d}g_a}{\prod_{k \in \lcb 1, 2, \dots, d\rcb}\partial z_k}(\bm z)\prod_{k \in \lcb 1, 2, \dots, d\rcb}\sign( x_k - z_k) \intd z_1, \dots \intd z_d
	\\ &  = 2 \cdot \int_{\realset^{d-1}}  \frac  {\partial ^{d-1}g_a}{\prod_{k \in \lcb 2, \dots, d\rcb}\partial z_k}(x_1,z_2,\dots,z_d)\prod_{k \in \lcb 2, \dots, d\rcb}\sign( x_k - z_k) \intd z_2, \dots \intd z_d
	\\ & = \dots
	\\ &  = 2^d \cdot  g_a( x_1, \dots, x_d)
	\\ &  = 2^d \cdot f(x_1,\dots, x_d) \cdot \exp(-\frac{\norm{\bm x-\bm x}_1}{a})
	\\ &  = 2^d \cdot f(\bm x). \label{integration_rn}
\end{align}
Combining \eqref{derivation_rn} and \eqref{integration_rn} completes the proof.
\end{proof}

\subsubsection{Compact Space \texorpdfstring{$\left[ 0 , 1 \right]^d$}{}}
\label{proof:multivariate:compact}
We now restate \sref{Lemma}{lemma:compactintegralform:main} in manner which facilitates the subsequent proof of the multivariate case.
\begin{lemma}
\label{lem:univariate:compact}
Let $f:(-1,+1)\rightarrow \realset$, $a>0$, and define
\begin{align}
	k^{\text{lr}}_a(\xl, \xr) 
	& \equiv
	\frac{\cosh((1+\xl)/a) \cosh((1-\xr)/a)}{a \sinh(2/a)}
	\\
	k_a(z,x) 
	& \equiv
	k^{\text{lr}}_a(\min(x,z), \max(x,z))
	\\
	r_a(z,x)
	& \equiv
	\big(1+\sign(x-z)\tanh((1+\sign(x-z)z)/a)\big)
	\\
	\kappa_a(z,x)
	& \equiv 
	r_a(z,x)k_a(z,x)
	\\
	b(a,x,f)
	& \equiv 
	\frac{\cosh(\frac{1-x}{a})f(-1)-\cosh(\frac{1+x}{a})f(1)}{\sinh(2/a)}.
\end{align}
Then $f$ has the integral representation
\begin{align}	
	f(x) = \int_{-1}^{+1}(f(z)+a f'(z))\kappa_a(z,x) \intd z+b(a,x,f).
\end{align}
\end{lemma}
Now for the multivariate version \sref{Lemma}{lem:multivariate:compact} we have the following
\begin{proof}
Let $f:[-1,1]^d\rightarrow \realset$ be $d$-times differentiable and $a>0$. Let define for any $\bm x$ in $[-1,1]^d$ and $\bm z$ in $[-1,1]^d\setminus\lbrace{\bm x}\rbrace$: 
\begin{align}
   g^i_a(\bm z,\bm x) & \equiv f(z_1, \dots ,z_d)\cdot\kappa_a(z_i,x_i),
\shortintertext{and}
    g_a(\bm z, \bm x) & \equiv f(z_1, \dots ,z_d)\cdot \prod_{i=1}^d  \kappa_a(z_i,x_i).
\end{align}
The idea for the compact domain is similar to the full Euclidean case. We will compute for $a >0$ and $\bm x \in [-1,1]^d$, $I(a,\bm x) $ in two different ways,
where
\begin{align}
I(a,\bm x)  = \int_{[-1,1]^d}   \frac  {\partial ^{d}g_a}{\prod_{k \in \lcb 1, 2, \dots, d\rcb}\partial z_k}(\bm z, \bm x) \intd  \bm z.
\end{align}
We will integrate the expression successively and then we will compute the derivative of the expression inside the integral.  

Regarding the integration, we need some preliminary results with respect to the different values of $\kappa_a$ defined in the lemma. 
\begin{lemma} 
The following properties are true for any $x \in [-1,1]$:
\begin{align}
    \kappa_a(1,x) & = \frac{B_a(x)}{a} \label{kappa_+1} \\
    \kappa_a(-1,x) & = \frac{B_a(-x)}{a} \label{kappa_-1} \\
    \kappa_a(x^-,x) - \kappa_a(x^+,x) & = \frac{1}{a} \label{Dkappa}.
\end{align}
\end {lemma}
\begin{proof}
Let's remind us the expressions of $k_a,r_a,\kappa_a$ for $x \in [-1,1]$ and $z \in [-1,1]\setminus\lbrace{x}\rbrace$: 
\begin{align}
    k_a(z,x) & = \frac{\cosh((1+\min(x,z))/a) \cosh((1-\max(x,z))/a)}{a \sinh(2/a)} 
    \\ & = \frac{\cosh(\frac{1+\sign(x-z)z}{a}) \cosh(\frac{1-\sign(x-z)x}{a})}{a \sinh(\frac{2}{a})}
    \\
    r_a(z,x) & = \big(1+\sign(x-z)\tanh((1+\sign(x-z)z)/a)\big)
    \\
    \kappa_a(z,x) & = r_a(z,x)\cdot k_a(z,x).
\end{align}
\subparagraph{When $z < x$:}
We get: $r_a(z,x) = \frac{1+\tanh(1+z)}{a}$ and $ k_a(z,x) = \frac{\cosh((\frac{1+z}{a}) \cosh(\frac{1-x}{a})}{a \sinh(2/a)}$ then: 
\begin{align}
	\kappa_a(-1,x) & = r_a(-1,x)\cdot k_a(-1,x) = 1\cdot\frac{\cosh(\frac{12x}{a})}{a\sinh(\frac{2}{a})}  =\frac{B_A(-x)}{a}.
\shortintertext{which proves \eqref{kappa_-1}, and}
	\kappa_a(x^-,x) & = r_a(x^-,x)\cdot k_a(x^-,x) 
    \\ & = \frac{\cosh(\frac{1+x}{a})-\sinh(\frac{1+x}{a})}{\cosh(\frac{1+x}{a})} \cdot \frac{\cosh(\frac{1+x}{a})\cdot\cosh(\frac{1-x}{a})}{a\sinh(\frac{2}{a})}
    \\ & = \frac{\cosh(\frac{1-x}{a})}{a\cdot\sinh(\frac{2}{a})}\big[ \cosh(\frac{1+x}{a})+\sinh(\frac{1+x}{a})\big] \label{kappa_x-}.
\end{align}
\subparagraph{When $z > x$:}
We get: $r_a(z,x) = \frac{1-\tanh(1-z)}{a}$ and $ k_a(z,x) = \frac{\cosh(\frac{1+x}{a}) \cosh(\frac{1-z}{a})}{a \sinh(2/a)}$ then:
\begin{align}
 \kappa_a(1,x) & = r_a(1,x)\cdot k_a(1,x) = 1\cdot\frac{\cosh(\frac{1+x}{a})}{a\sinh(\frac{2}{a})}  =\frac{B_a(x)}{a},
\shortintertext{which proves \eqref{kappa_+1}, and}
    \kappa_a(x^+,x) & = r_a(x^+,x)\cdot k_a(x^+,x) 
    \\ & = \frac{\cosh(\frac{1-x}{a})-\sinh(\frac{1-x}{a})}{\cosh(\frac{1-x}{a})} \cdot \frac{\cosh(\frac{1+x}{a})\cdot\cosh(\frac{1-x}{a})}{a\sinh(\frac{2}{a})} 
    \\ & = \frac{\cosh(\frac{1+x}{a})}{a\cdot\sinh(\frac{2}{a})}\big[ \cosh(\frac{1-x}{a})-\sinh(\frac{1-x}{a})\big] \label{kappa_x+}.
\end{align}
Finally with $\eqref{kappa_x-}$ and $\eqref{kappa_x+}$, we can prove $\eqref{Dkappa}$:
\begin{align}
    \kappa_a(x^-,x) - \kappa_a(x^+,x) & = \frac{\cosh(\frac{1-x}{a})}{a\cdot\sinh(\frac{2}{a})}\big[ \cosh(\frac{1+x}{a}) + \sinh(\frac{1+x}{a})\big] 
    \\ & ~~~~~ - \frac{\cosh(\frac{1+x}{a})}{a\cdot\sinh(\frac{2}{a})}\big[ \cosh(\frac{1-x}{a}) - \sinh(\frac{1-x}{a})\big]
    \\ & = \frac{\cosh(\frac{1-x}{a})  \cosh(\frac{1+x}{a}) +  \cosh(\frac{1-x}{a}) \sinh(\frac{1+x}{a})  }{a\cdot\sinh(\frac{2}{a})} 
    \\ & ~~~~~ - \frac{ \cosh(\frac{1+x}{a})  \cosh(\frac{1-x}{a})  -   \cosh(\frac{1+x}{a})  \sinh(\frac{1-x}{a}) }{a\cdot\sinh(\frac{2}{a})}
    \\ & = \frac{ \cosh(\frac{1-x}{a}) \sinh(\frac{1+x}{a}) +  \cosh(\frac{1+x}{a})  \sinh(\frac{1-x}{a}) }{a\cdot\sinh(\frac{2}{a})}
    \\ & = \frac{\sinh(\frac{2}{a})}{a\cdot\sinh(\frac{2}{a})}
    \\ & = \frac{1}{a}.
\end{align}
\end{proof}
We can now get back to $I(a,\bm x)$ and integrate the expression successively: 
\subparagraph{Step 1: Integration}
Let's prove the integration for one integral and then let's repeat the integration successively by recurrence. For $i$ in $\lcb 1, 2, \dots, d\rcb$, and $\bm x, \bm z $ in $ [-1,1]^{d}$:

To compute the integral, we have to pay attention that $x_i$ is a point of discontinuity of $\big( t \to \kappa_a(t,x_i )\big)$, we need to cut the integral at this point.
\newcommand\sss{\hspace{4.4mm}}
\begin{align}
     & \int_{[-1,1]} \frac  {\partial  g^i_a }{\partial z_i}(\bm z,\bm x)\intd z_i 
    \\ & =\int_{[-1,1]} \frac  {\partial f(\bm z)\cdot\kappa_a(z_i,x_i)}{\partial z_i}\intd z_i 
    \\
    & = \int_{[-1,x_i[ }\frac  {\partial f(\bm z)\cdot\kappa_a(z_i,x_i)}{\partial z_i}\intd z_i 
     +\int_{]x_i,1]} \frac  {\partial f(\bm z)\cdot\kappa_a(z_i,x_i)}{\partial z_i}\intd z_i.
     \\ & = \Bigg[  f(\bm z)\cdot\kappa_a(z_i,x_i)   \Bigg]_{x_i^+}^{1} + \Bigg[  f(\bm z)\cdot\kappa_a(z_i,x_i)   \Bigg]_{-1}^{x_i^-}
    \\
    & = \bigg[ f(z_1, \dots z_{i-1},1,z_{i+1}, \dots ,z_d) \cdot  \kappa_a(1,x_i) 
      - f(z_1, \dots z_{i-1},x_i^+,z_{i+1}, \dots ,z_d) \cdot  \kappa_a(x_i^+,x_i) \bigg]
    \\ & ~~ +
    \bigg[ f(z_1, \dots z_{i-1},x_i^-,z_{i+1}, \dots ,z_d) \cdot  \kappa_a(x_i^-,x_i) 
     - f(z_1, \dots z_{i-1},-1,z_{i+1}, \dots ,z_d) \cdot  \kappa_a(-1,x_i) \bigg]
    \\
     & = \Bigg[ f(z_1, \dots z_{i-1},1,z_{i+1}, \dots ,z_d) \cdot  \frac{B_a(x_i)}{a}
     \sss + \frac{f(z_1, \dots z_{i-1},x_i,z_{i+1}, \dots ,z_d)}{a} 
    \\ & \sss  - f(z_1, \dots z_{i-1},-1,z_{i+1}, \dots ,z_d) \cdot  \frac{B_a(-x_i)}{a} \Bigg]
    \\
    & = \frac{ \sum_{\substack{\epsilon_i \in \lcb -1, 0, 1\rcb}} B^{ \epsilon_i}(x_i) \cdot f(z_1, \dots z_{i-1},x^{\epsilon_i}_i,z_{i+1},
    \dots ,z_d)}{a},
\end{align}
  where $B^{ \epsilon_i}(x_i) =  B(\epsilon_i x_i)\epsilon_i + (1-\epsilon_i)(1+\epsilon_i)$ and $x^{\epsilon_i}_i = \epsilon + (1-\epsilon)(1+\epsilon) x_i$.

This latter notation of  $B^{ \epsilon_i}(x_i)$ and $x^{\epsilon_i}_i$ enable a clear expression of the integral, and we find the different terms immediately:
\begin{alignat}{3}
      &\text{if } \epsilon_i =-1 \text{ : } && B_a^{ \epsilon_i}(x_i) =  -B_a(-x_i)& & \text{ and }  x^{\epsilon_i}_i  = - 1 \\
      &\text{if } \epsilon_i  = 0 \text{ : } && B_a^{ \epsilon_i}(x_i) = 1& & \text{ and }  x^{\epsilon_i}_i  = x_i \\
      &\text{if }  \epsilon_i  = 1 \text{ : } && B_a^{ \epsilon_i}(x_i) = B_a(x_i) & & \text{ and } x^{\epsilon_i}_i  = 1 \\
\end{alignat}
Then we proceed by recurrence for $I(a,\bm x)$:
\begin{align}    
	I(a,\bm x) &  =
      \int_{[-1,1]^d}  
	 \frac  {\partial ^{d}g_a}{\prod_{k \in \lcb 1, 2, \dots, d\rcb}\partial z_k}(\bm z, \bm x)  \intd \bm z
	 \\ & =  \int_{[-1,1]^d}  
	 \frac  {\partial ^{d}f(z_1, \dots ,z_d)\cdot \prod_{i=1}^d  \kappa_a(z_i,x_i)}{\prod_{k \in \lcb 1, 2, \dots, d\rcb}\partial z_k}(\bm z, \bm x) \intd z_1\dots  \intd z_d
	\\ & =  \int_{[-1,1]^{d-1}} \int_{[-1,1]}  \frac{\partial}{\partial z_1} \Bigg( \frac  {\partial ^{d-1} f(z_1, \dots ,z_d)\cdot \prod_{i=2}^d  \kappa_a(z_i,x_i) }{\prod_{k \in \lcb  2, \dots, d\rcb}\partial z_k} \kappa_a(z_1,x_1) \Bigg) \intd z_1\dots  \intd z_d
	\\ & =  \int_{[-1,1]^{d-1}}  \frac  {\sum_{\substack{\epsilon_i \in \lcb -1, 0, 1\rcb}} B_a^{ \epsilon_1}(x_1) \cdot \frac  {\partial ^{d-1} 		f(x^{\epsilon_1}_1,z_2, \dots ,z_d)\cdot \prod_{i=2}^d  \kappa_a(z_i,x_i) }{\prod_{k \in \lcb  2, \dots, d\rcb}\partial z_k}  }{a} \intd z_2 \dots  \intd z_d
	\\ & = \frac{ \sum_{\substack{\epsilon_i \in \lcb -1, 0, 1\rcb}} B_a^{ \epsilon_1}(x_1) \cdot \int_{[-1,1]^{d-1}} \frac  {\partial ^{d-1} 		f(x^{\epsilon_1}_1,z_2, \dots ,z_d)\cdot \prod_{i=2}^d  \kappa_a(z_i,x_i) }{\prod_{k \in \lcb  2, \dots, d\rcb}\partial z_k} \intd z_2 \dots  \intd z_d }{a}
  \intertext{and by recurrence}
	& = \cdots
   	\\ & = \frac{\sum_{\substack{\bm \epsilon \in \lcb -1, 0, 1\rcb^d} } \Big( \prod_{i \in \lcb 1, 2, \dots, d\rcb} B_a^{\epsilon_i}(x_i) \Big)\cdot f \big( x^{\epsilon_1}_1 ,\cdots, x^{\epsilon_d}_d \big) }{a^d}
	\\ & = \frac{\sum_{\substack{\bm \epsilon \in \{-1,0,1\}^d}} B_a^{\epsilon}(\bm x) \cdot f\big(\bm x^{\epsilon} \big) }{a^d}
	\\ & = \frac{f(\bm x) + \sum_{\substack{\bm \epsilon \in \{-1,0,1\}^d\setminus\lbrace{(0, \dots ,0)}\rbrace}} B_a^{\epsilon}(\bm x) \cdot f\big(\bm x^{\epsilon} \big) }{a^d} \label{step_1_compact}
\end{align}
We are now going to compute $I(a,\bm x)$  by differentiation. 
\subparagraph{Step 2: Differentiation}
First, let's compute the derivation of $\kappa_a$ for $x \in [-1,1]$ and $z \in [-1,1]\setminus\lbrace{x}\rbrace$:  
\subparagraph{When $z < x$:}
\begin{align}
	 \frac{ \partial r_a }{ \partial z }(z,x) & = \frac{1-\tanh^2(\frac{1+z}{a})}{a} =\frac{1-(1-r_a(z,x))^2}{a} 
 \shortintertext{and}
 	\frac{\partial k_a}{\partial z}(z,x) & = \frac{\sinh((\frac{1+z}{a}) \cosh(\frac{1-x}{a})}{a^2 \sinh(2/a)}
\end{align}
\subparagraph{When $z > x$:}
\begin{align}
 	\frac{\partial r_a}{\partial z}(z,x) & =\frac{1-\tanh^2(\frac{1-z}{a})}{a} =\frac{1-(1-r_a(z,x))^2}{a} 
 \shortintertext{and}
 	\frac{\partial k_a}{\partial z}(z,x) & = -\frac{\cosh(\frac{1+x}{a}) \sinh(\frac{1-z}{a})}{a^2 \sinh(2/a)} 
\end{align}
\subparagraph{Finally when $z \in [-1,1]\setminus\lbrace{x}\rbrace$:}
\begin{align}
	\frac{\partial r_a}{\partial z}(z,x) & =\frac{1-(1-r_a(z,x))^2}{a} 
 \shortintertext{and}
 	\frac{\partial k_a}{\partial z}(z,x) & = \sign(x-z)\frac{\cosh(\frac{1-\sign(x-z)\cdot x}{a}) \sinh(\frac{1+\sign(x-z)\cdot z}{a})}{a^2 \sinh(2/a)} 
    \\ & = \frac{\sign(x-z) \sinh(\frac{1+\sign(x-z)\cdot z}{a}) }{a \cosh(\frac{1+\sign(x-z)\cdot z}{a}) } \frac{\cosh(\frac{1-\sign(x-z)\cdot x}{a}) \cdot \cosh(\frac{1+\sign(x-z)\cdot z}{a})}{ a \sinh(\frac{2}{a})  } 
    \\ & = \frac{ \sign(x-z) \tanh(\frac{1+\sign(x-z)\cdot z}{a})}{a}\cdot k_a(z,x) 
    \\ & = \frac{(r(z,x)-1)\cdot(k(z,x))}{a}. \label{dk_dz}
\shortintertext{Therefore}  
	\frac{\partial \kappa_a}{\partial z}(z,x) & = \big(\frac{\partial r_a}{\partial z}\cdot k_a + \frac{\partial k_a}{\partial z}\cdot r_a \big) (z,x)
    \\ & = \frac{1-(1-\cdot r_a(z,x))^2}{a} \cdot k_a(z,x) + \frac{(r_a(z,x)-1)\cdot(k_a(z,x))}{a} \cdot r_a(z,x) 
    \\ & = \frac{2\cdot r_a(z,x)k_a(z,x) - r_a^2(z,x)\cdot k_a(z,x)}{a} 
    \\ & + \frac{r_a^2(z,x)\cdot k_a(z,x)- k_a(z,x)\cdot r_a(z,x) }{a} 
    \\ & = \frac{r_a(z,x)\cdot k_a(z,x)}{a}
    \\ & = \frac{\kappa_a(z,x)}{a}.
\end{align}
We can now compute the derivative of $g$ inside $I$ for $\bm x$ in $ [-1,1]^{d}$ and $ \bm z $ in $ [-1,1]^{d}\setminus\lbrace{\bm x}\rbrace$
Using the formula of the product derivative,
\begin{align}
    \frac  {\partial ^{d}g_a}{\prod_{k \in \lcb 1, 2, \dots, d\rcb}\partial z_k}(\bm z, \bm x)
    & = \sum_{\substack{\mathcal J \subseteq \lcb 1, 2, \dots, d\rcb \\ }} \frac  {\partial ^{\abs{\mathcal{J}}}f}{\prod_{k \in \mathcal J}\partial z_k}(\bm z) \cdot \frac  {\partial ^{d - \abs{\mathcal{J}}}\prod_{i=1}^d  \kappa_a(z_i,x_i)}{\prod_{k \notin \mathcal J}\partial z_k} 
    \\ & = \sum_{\substack{\mathcal J \subseteq \lcb 1, 2, \dots, d\rcb \\ }} \frac  {\partial ^{\abs{\mathcal{J}}}f}{\prod_{k \in \mathcal J}\partial z_k} (\bm z)   \cdot  \frac{\prod_{i=1}^d  \kappa_a(z_i,x_i)}{a^{d - \abs{\mathcal{J}}}}
    \\ & = \frac{\sum_{\substack{\mathcal J \subseteq \lcb 1, 2, \dots, d\rcb \\ }} a^{\abs{\mathcal{J}}}\frac  {\partial ^{\abs{\mathcal{J}}}f}{\prod_{k \in \mathcal J}\partial z_k} (\bm z)\prod_{i=1}^d  \kappa_a(z_i,x_i)}{a^{d}} .
\end{align}
So if we integrate the expression above with respect to $\bm z$ in $[-1,1]^d$,
\begin{align}
    I(a,\bm x) & = \frac  {\int_{[-1,1]^d} \bigg( \sum_{\substack{\mathcal J \subseteq \lcb 1, 2, \dots, d\rcb \\ }} a^{\abs{\mathcal{J}}}\frac  {\partial ^{\abs{\mathcal{J}}}f}{\prod_{k \in \mathcal J}\partial z_k} (\bm z) \prod_{i=1}^d  \kappa_a(z_i,x_i) \bigg)\intd \bm z}{a^d}. \label{step_2_compact}
\end{align}
Combining \eqref{step_1_compact} and \eqref{step_2_compact} completes the proof.
\end{proof}

\section{Details of the Analytical Example \texorpdfstring{of \sref{Section}{sec:experiments:analytical}}{}}

We provide further analysis of the absolutely continuous component $\widehat p$ of the toy distribution $p$ of \autoref{sec:experiments:analytical}.

\label{sec:experiments:analytical:details}

\subsection{Theoretical Analysis}
\label{sec:experiments:analytical:theoretical}
\textit{The bold numbers below (approximately) match the corresponding bold values in \Autoref{fig:ft_demo_output}.}

Putting the definition \eqref{eqn:prational} into \textit{e.g.} the log-derivative trick yields in closed form
\begin{align}
\nabla_{\theta=0} \expect{x \sim p(\cdot|\theta)}{f(x)}
\equiv 
\widehat \mu
=
\frac{\pi -2}{\pi }
\approx
\pythonoutput{0.363}.
\end{align}
Take $n=2$ samples and define $x \equiv x^{(1)}$ and $z \equiv x^{(2)}$. The variance of the log-derivative estimator \eqref{eqn:logtrick:sum} for $\widehat \mu$ at $\theta=0$ is then, defining $g_{\text{ld}}(x)\equiv f(x)\nabla_\theta \log p(x|\theta)$,
\begin{align}
V_\mathrm{ld} 
 \equiv 
\expectlr{x,z\sim p(\cdot|\theta=0)}{\lnb \frac{1}{2}\big( g_{\text{ld}}(x) + g_{\text{ld}}(z)\big)-\mu\rnb^2 }
 = \frac{1}{4}-\frac{2}{\pi ^2}
\approx
\pythonoutput{0.0474}.
\label{eqn:vld}
\end{align}
The corresponding variance of our new estimator \eqref{eqn:logreparameterisationtrick:scalar:doublesum} (for simplicity we avoid the further symmetrised \eqref{eqn:logreparameterisationtrick:scalar:doublesum}) is, defining $g_\mathrm{ft}(x,z)=\frac{f'(z) \frac{1}{2} \sign (z-x)}{p(z|\theta)} \nabla_\theta \log p(x|\theta)$,
\begin{align}
V_\mathrm{ft}  \equiv \expectlr{x,z\sim p(\cdot|\theta=0)}{\big( g_\mathrm{ft}(x,z)-\mu\big)^2}
 = \frac{\pi }{12}-\frac{(\pi -2)^2}{\pi ^2}
\approx
\pythonoutput{0.130}.
\label{eqn:vft}
\end{align}
This is worse than $V_\mathrm{ld}$; but the convex combination 
\begin{align}
& g_{\text{ld+ft}}(x,z,c)
=
c \times \frac{1}{2}\big( g_{\text{ld}}(x) + g_{\text{ld}}(z)\big) + (1-c) \times g_\mathrm{ft}(x,z)
\end{align}
retains the correct expectation while enjoying a variance of
\begin{align}
& V_\mathrm{ld+ft} 
\equiv 
\expectlr{x,z\sim p(\cdot|\theta=0)}{\lnb g_\mathrm{ld+ft}(x,z,c)-\mu\rnb^2}
\\ & ~~= \frac{1}{12\pi^2}
\Big(\left(24+72 \pi -27 \pi ^2+\pi ^3\right) c^2+\pi ^3-12 \pi ^2 
+48 \pi -48 -2 \pi  \left(60-21 \
\pi +\pi ^2\right) c\Big).
\end{align}
This is minimised at (it is straightforward to check that the same result may also be obtained in terms of the uncentered moments using \eqref{eqn:cstar})
\begin{align}
\label{eqn:theoreticalc}
c^\star=\frac{\pi  \left(60-21 \pi +\pi ^2\right)}{24+72 \pi -27 \pi ^2+\pi ^3}\approx \pythonoutput{0.831},
\end{align} 
obtaining a variance of
\begin{align}
& V_\mathrm{ld+ft}^\star \equiv \expectlr{x,z\sim p(\cdot|\theta=0)}{\lnb g_\mathrm{ld+ft}(x,z,c^\star)-\mu\rnb^2}
\\ & ~~= \frac{-384-768 \pi +288 \pi ^2+112 \pi ^3-39 \pi ^4+\pi ^5}{4 \pi ^2 \left(24+72 \pi -27 \pi ^2+\pi ^3\right)}
\approx \pythonoutput{0.925} \times V_\mathrm{ld},
\label{eqn:theoreticalvarianceratio}
\end{align}
where $V_\mathrm{ld}$ is the log-derivative estimator variance of \eqref{eqn:vld}.

\subparagraph{Baseline}

One typically includes a variance reducing baseline in the log-derivative estimator, replacing $g_\mathrm{ld}(x)$ with the still unbiased 
\begin{align}
\label{eqn:baseline}
g_{\text{ld+b}}(x,b)\equiv (f(x)-b)\nabla_\theta \log p(x|\theta)
\end{align}
(see \textit{e.g.} \citet{petersbaseline} for a detailed discussion of this approach). %
It is easy to verify however, that for this example the optimal baseline is $b=0$.

\subsection{Further Details on the Numerical Toy Experiments}
\label{sec:experiments:numerical}

\subparagraph{Approximation of $c^\star$.} 

In practice the optimal mixing proportion $c^\star$ will be unknown. We handle this by simply estimating the required moments (three real numbers: two variances and a covariance) for \eqref{eqn:cstar} empirically based on the given samples $x^{(i)}$ --- for a more general discussion of estimator combination see \textit{e.g.} \citet{domkecontrolvariates}. In our numerical experiments we use \textit{centered} empirical second moments (\textit{i.e.} covariances), as this guarantees that the variance is minimised for the given empirical sample (the result \eqref{eqn:cstar} uses the fact that the true expectations match). Our approach is demonstrated in the included python script (mentioned at the top of \Autoref{sec:experiments:analytical}), and obtains a close match to the theoretical results above in terms of both $c^\star$ and (most importantly) the obtained variance reduction --- see \Autoref{fig:ft_demo_output}.

\subparagraph{The power of pairwise interactions.}

The analytical results above show that our new fundamental trick estimator yields a reduction in variance, but only in convex combination with the standard log-derivative estimator. While this is already promising, an important phenomenon arises as we increase the number of samples beyond the minimal $n=2$ considered above. This is illustrated in \Autoref{fig:symdemo}, the value at $n=2$ for the blue (orange) line of which matches the theoretical value of \eqref{eqn:vld} (\eqref{eqn:vft}). While the variance of the log-derivative estimator obviously decreases linearly in $n$, the double summation of the fundamental trick estimator leads to a favourable scaling, especially for small $n$. Remarkably, our new estimator rapidly achieves a superior variance all on its own, without requiring convex combination as above. This phenomenon is especially interesting given the \textit{minibatch} size in stochastic gradient descent is typically moderate, say $n\approx 10 \text{~to~} 100$, in size \citep{hinton_cookbook}.

\subparagraph{Issues arising from the $\frac{1}{p(x|\theta)}$ factor.} 

\begin{figure}[ht]%
  \begin{center}%
    \includegraphics[page=1,height=0.3\textwidth]{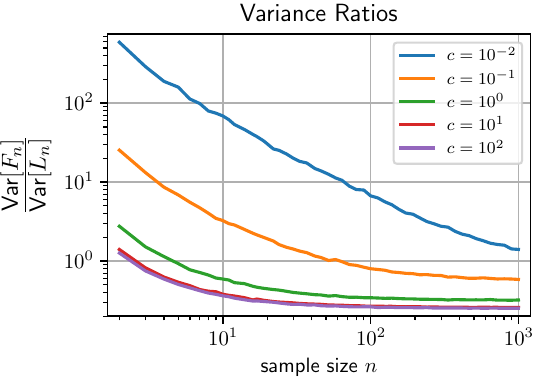}
    ~\\~\\
    \includegraphics[page=1,height=0.32\textwidth]{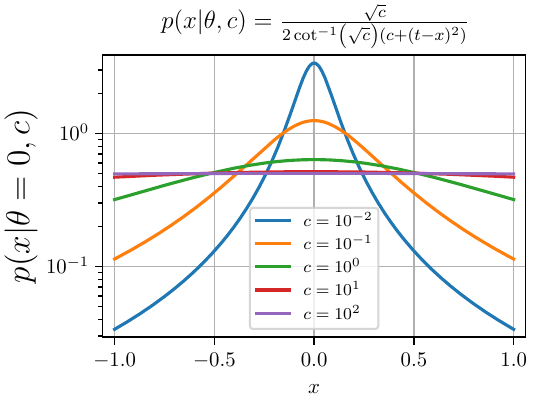}
  \end{center}%
  \caption{%
    On the top we plot ratios of the two variances previously plotted in \autoref{fig:symdemo} as a function of the number of samples $n$ --- see the caption of that figure for more details. Here we vary the parameter $c$ in the probability distribution $p(x|\theta)\propto 1/(c+(x-\theta)^2)$, which is depicted on the lower figure. Hence the upper $c=1$ curve corresponds to (the ratio of the two curves plotted in) \autoref{fig:symdemo}. As expected, we observe that the variance of the fundamental trick increases as the minimum of the distribution approaches zero (\ie\ as $c\rightarrow 0$).
    \label{fig:sym_demo_denominatorissue}%
  }%
\end{figure}

All of our new pairwise estimators sum terms with a probability in the denominator. The supplementary \Autoref{fig:sym_demo_denominatorissue} explores the effect this has, by varying $c$ (which is $1$ in \eqref{eqn:prational} above) in the more general $p(x|\theta)\propto 1/(c+(x-\theta)^2)$. As expected, the fundamental trick estimator performs better for larger $c$, as the denominator is sufficiently large. While it is tempting to eliminate this term as in the supplementary \sref{Appendix}{sec:hsl}, we show that this results in a non-diversifying estimator. A similar issues arises in importance sampling, which involves a ratio of probabilities, and the recent work addressing this issue in that case may be appropriate here \citep{ionides2008,anuinfinitevariance}, but is beyond the present scope.

\subparagraph{The power of the representer trick.} 

\begin{figure*}[t]%
  \hfill\hfill
  \includegraphics[page=2,width=0.44\textwidth]{figs/cov_demo_1/cov_demo_fig.pdf}
  ~~~~~
  \includegraphics[page=1,width=0.44\textwidth]{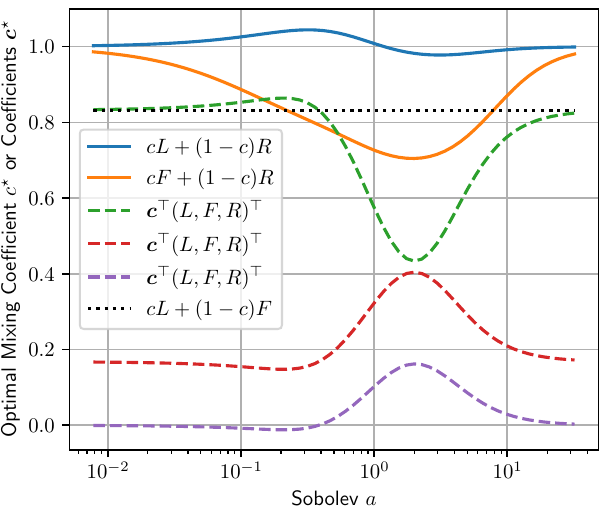}
  \hfill
  ~\\
  \caption{
  \label{fig:covdemo}
  Convex combinations of estimators on the toy problem of \autoref{sec:experiments}, for the minimal $n=2$ samples. We plot variance ratios (left) and optimal mixing coefficients (right) for various convex combinations of the $L$ (log-derivative), $F$ (fundamental trick) and $R$ (representer trick) estimators, as a function of the $a$ parameter of the compact Sobolev space of \autoref{sec:sobolev} (here this parameter affects $R$). On the r.h.s. plot, for mixtures of just two estimators we plot the scalar $c$ as the other coefficient is simply $1-c$, while for the combination of all three estimators we plot three dashed green, red, and purple lines corresponding to the $L$, $F$ and $R$ components of the vector of coefficients $\bm c$. The dotted black line on the r.h.s. (respectively l.h.s.) plot is flat as it does not depend on $c$, and matches the theoretical value of  \eqref{eqn:theoreticalc} (respectively \eqref{eqn:theoreticalvarianceratio}). We observe that the lowest overall variance corresponds to the minima of the green l.h.s. curve, whereupon the variance is reduced by slightly more than 20\% w.r.t. the standard log-derivative estimator.
  }
\end{figure*}

We extended the convex combination of fundamental trick and log-derivative trick to include our more advanced representer trick, via the minimum variance convex combination over all three estimators --- see \Autoref{fig:covdemo}. The result is a scheme which (here again for $n=2$ samples) further reduces the variance to less than $80\%$ of that of the log-derivative estimator. To reveal the true power of the representer trick, however, we  experiment with non-linear test functions $f(x)=\sin(\pi\omega x)$, for $\omega\in\{1,2,3\}$, in the supplementary \Autoref{fig:cov_demo_extra}. As expected, while the non-local nature of the fundamental trick leads to a breakdown for larger $\omega$, the representer trick merely requires a matching length scale parameter $a$ for the Sobolev space of \Autoref{sec:sobolev}. Note that in these cases again, due to symmetry the optimal variance reducing baseline of \eqref{eqn:baseline} is zero.

\clearpage

\begin{figure}
\hspace{25mm}
{
\texttt{
\footnotesize
\minibox[frame]{
x=log-derivative, L: \\
 E[x] = \pythonoutput{0.3633} E[x\textasciicircum 2] = 0.1794 VAR[x] = \pythonoutput{0.0474} \\
x=fundamental, F: \\
 E[x] = \pythonoutput{0.3635} E[x\textasciicircum 2] = 0.2619 VAR[x] = \pythonoutput{0.1297} \\
x=representer a = 0.20, R: \\
 E[x] = \pythonoutput{0.3676} E[x\textasciicircum 2] = 1.3989 VAR[x] = 1.2638 \\
x=optimal cL + (1-c)F: \\
 E[x] = \pythonoutput{0.3633} E[x\textasciicircum 2] = 0.1759 VAR[x] = 0.0439 \\
optimal c in cL + (1-c)F = \pythonoutput{0.8310} \\
VAR[optimal cL + (1-c)F] / VAR[L] = \pythonoutput{0.9251} \\
x=optimal cL + (1-c)R: \\ 
 E[x] = \pythonoutput{0.3631} E[x\textasciicircum 2] = 0.1774 VAR[x] = 0.0456 \\
optimal c in cL + (1-c)R = \pythonoutputb{1.0399} \\
VAR[optimal cL + (1-c)R] / VAR[L] = \pythonoutputb{0.9621} \\
}
}
~\\~\\
}
\caption{
\label{fig:ft_demo_output}
Output of the Supplementary Demo Program \texttt{gradtricks\_demo.py}
	The bold numbers above (approximately) match the corresponding bold values in \sref{Appendix}{sec:experiments:analytical:theoretical}. The italic bold numbers pertain to the optimal combination of the representer trick (with $a=0.2$) and the log-derivative trick, and may also be read off \sref{Figure}{fig:covdemo}.
}
\end{figure}

\newcommand\onefig[2]{%
~\\~\\~\\
  \subfigure[#2]{%
    \includegraphics[page=2,height=0.28\textwidth]{figs/cov_demo_#1.pdf}%
    ~~
    \includegraphics[page=3,height=0.28\textwidth]{figs/cov_demo_#1.pdf}
  }%
}
\begin{figure*}[ht]%
  \begin{center}
  \onefig{sina}{$f(x)=\sin(\pi x)$}
  \onefig{sinb}{$f(x)=\sin(2\pi x)$}
  \onefig{sinc}{$f(x)=\sin(3\pi x)$}
  \end{center}
  \caption{
    Variance ratios (left) for various test functions (right) of increasing non-linearity (top to bottom). The l.h.s. figures above are similar to the l.h.s. figure of \autoref{fig:covdemo} --- see the corresponding caption for details. We observe that the fundamental trick breaks for overly non-linear functions (as evidenced by the dotted black line on the l.h.s. figures) while the representer trick does not. As expected, for increasingly non-linear test functions, the optimal length scale $a$ of the Sobolev space decreases.
    \label{fig:cov_demo_extra}
  }
\end{figure*}

\end{document}